\documentclass[10pt,twocolumn,letterpaper]{article}

\usepackage{cvpr}
\usepackage{times}
\usepackage{epsfig}
\usepackage{graphicx}
\usepackage{amsmath}
\usepackage{amssymb}

\graphicspath{{figures/},{figures/n14/images/},{figures/n14-vs-n1-to-zero/images/},{figures/train-stat/}}

\usepackage[pagebackref=true,breaklinks=true,letterpaper=true,colorlinks,bookmarks=false]{hyperref}

\cvprfinalcopy % *** Uncomment this line for the final submission

\def\cvprPaperID{xxxx} % *** Enter the CVPR Paper ID here

\begin{document}

%%%%%%%%% TITLE
\title{Fast-SCNN: Fast Semantic Segmentation Network}

\author{Rudra PK Poudel\\
Toshiba Research Europe, UK\\
{\tt\small rudra.poudel@crl.toshiba.co.uk}
% For a paper whose authors are all at the same institution,
% omit the following lines up until the closing ``}''.
% Additional authors and addresses can be added with ``\and'',
% just like the second author.
% To save space, use either the email address or home page, not both
\and Stephan Liwicki\\
Toshiba Research Europe, UK\\
{\tt\small stephan.liwicki@crl.toshiba.co.uk}
\and
Roberto Cipolla\\
Cambridge University, UK\\
{\tt\small rc10001@cam.ac.uk}
}

\maketitle
%\thispagestyle{empty}

%%%%%%%%% ABSTRACT
\begin{abstract}
The encoder-decoder framework is state-of-the-art for offline semantic image segmentation. Since the rise in autonomous systems, real-time computation is increasingly desirable. In this paper, we introduce fast segmentation convolutional neural network (Fast-SCNN), an above real-time semantic segmentation model on high resolution image data ($1024\times2048px$) suited to efficient computation on embedded devices with low memory. Building on existing two-branch methods for fast segmentation, we introduce our `learning to downsample' module which computes low-level features for multiple resolution branches simultaneously. Our network combines spatial detail at high resolution with deep features extracted at lower resolution, yielding an accuracy of 68.0\% mean intersection over union at 123.5 frames per second on Cityscapes. We also show that large scale pre-training is unnecessary. We thoroughly validate our metric in experiments with ImageNet pre-training and the coarse labeled data of Cityscapes. Finally, we show even faster computation with competitive results on subsampled inputs, without any network modifications.
\end{abstract} 

%%%%%%%%% BODY TEXT
\section{Introduction}
Fast semantic segmentation is particular important in real-time applications, where input is to be parsed quickly to facilitate responsive interactivity with the environment. Due to the increasing interest in autonomous systems and robotics, it is therefore evident that the research into real-time semantic segmentation has recently enjoyed significant gain in popularity \cite{contextnet-poudel2018,BiSeNet-yu2018,gun-mazzini2018,erfnet-romera2018,icnet-zhao2017b,enet-paszke2016}. We emphasize, faster than real-time performance is in fact often necessary, since semantic labeling is usually employed only as preprocessing step of other time-critical tasks. Furthermore, real-time semantic segmentation on embedded devices (without access to powerful GPUs) may enable many additional applications, such as augmented reality for wearables.

We observe, in literature semantic segmentation is typically addressed by a deep convolutional neural network (DCNN) with an encoder-decoder framework \cite{fcn-long2016,segnet-badrinarayanan2017}, while many runtime efficient implementations employ a two- or multi-branch architecture \cite{contextnet-poudel2018,BiSeNet-yu2018,gun-mazzini2018}. It is often the case that
\begin{itemize}
  \item{a larger receptive field is important to learn complex correlations among object classes (\ie global context),}
  \item{spatial detail in images is necessary to preserve object boundaries, and}
  \item{specific designs are needed to balance speed and accuracy (rather than re-targeting classification DCNNs).}
\end{itemize}
Specifically in the two-branch networks, a deeper branch is employed at low resolution to capture global context, while a shallow branch is setup to learn spatial details at full input resolution. The final semantic segmentation result is then provided by merging the two. Importantly since the computational cost of deeper networks is overcome with a small input size, and execution on full resolution is only employed for few layers, real-time performance is possible on modern GPUs. In contrast to the encoder-decoder framework, the initial convolutions at different resolutions are not shared in the two-branch approach. Here it is worth noting, the guided upsampling network (GUN)~\cite{gun-mazzini2018} and the image cascade network (ICNet)~\cite{icnet-zhao2017b} only share the weights among the first few layers, but not the computation.

\begin{figure*}[t]
\begin{center}
   \includegraphics[width=1\linewidth]{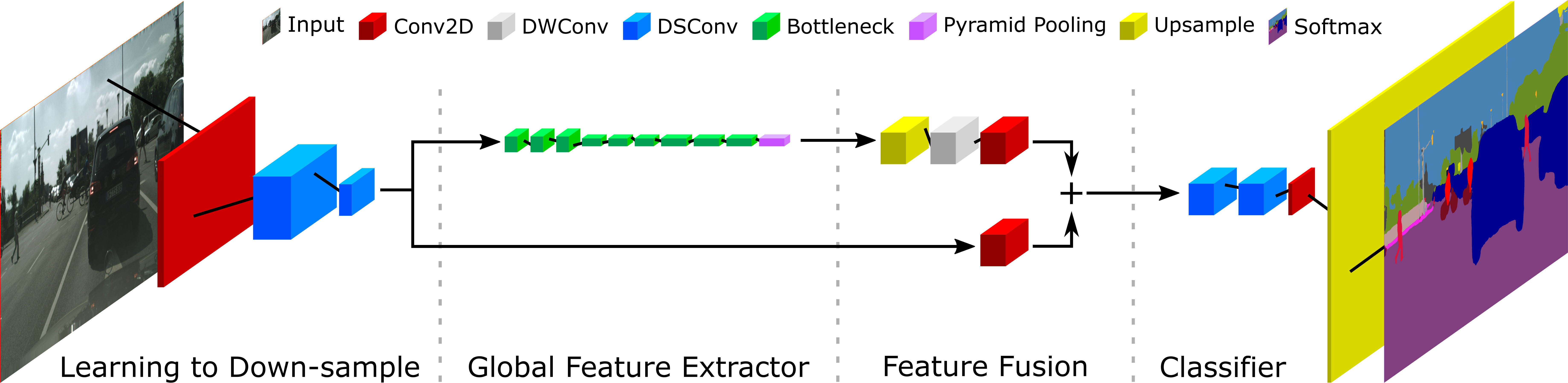}
\end{center}
\caption{Fast-SCNN shares the computations between two branches (encoder) to build a above real-time semantic segmentation network.}
\label{fig:fast-scnn}
\end{figure*}
In this work we propose \textit{fast segmentation convolutional neural network} Fast-SCNN, an above real-time semantic segmentation algorithm merging the two-branch setup of prior art \cite{contextnet-poudel2018,BiSeNet-yu2018,gun-mazzini2018,icnet-zhao2017b}, with the classical encoder-decoder framework \cite{fcn-long2016,segnet-badrinarayanan2017} (Figure~\ref{fig:fast-scnn}). Building on the observation that initial DCNN layers extract low-level features \cite{deconv-zeiler2014,olah2017}, we share the computations of the initial layers in the two-branch approach. We call this technique \textit{learning to downsample}. The effect is similar to a skip connection in the encoder-decoder model, but the skip is only employed once to retain runtime efficiency, and the module is kept shallow to ensure validity of feature sharing. Finally, our Fast-SCNN adopts efficient depthwise separable convolutions \cite{depthwise-conv-sifre2014,mobilenet-howard2017}, and inverse residual blocks \cite{inverted-res-bottlenecks-sandler2018}.

Applied on Cityscapes \cite{cityscaples2016}, Fast-SCNN yields a mean intersection over union (mIoU) of 68.0\% at 123.5 frames per second (fps) on a modern GPU (Nvidia Titan Xp (Pascal)) using full ($1024 \times 2048px$) resolution, which is twice as fast as prior art i.e. BiSeNet (71.4\% mIoU)\cite{BiSeNet-yu2018}.

While we use 1.11 million parameters, most offline segmentation methods (\eg DeepLab \cite{deeplab-v2-chen2016} and PSPNet \cite{pspnet-zhao2017a}), and some real-time algorithms (\eg GUN \cite{gun-mazzini2018} and ICNet \cite{icnet-zhao2017b}) require much more than this. The model capacity of Fast-SCNN is kept specifically low. The reason is two-fold: (i) lower memory enables execution on embedded devices, and (ii) better generalisation is expected. In particular, pre-training on ImageNet \cite{imagenet2015} is frequently advised to boost accuracy and generality \cite{pspnet-zhao2017a}. In our work, we study the effect of pre-training on the low capacity Fast-SCNN. Contradicting the trend of high-capacity networks, we find that results only insignificantly improve with pre-training or additional coarsely labeled training data (+0.5\% mIoU on Cityscapes \cite{cityscaples2016}). In summary our contributions are:
\begin{enumerate}
\item We propose Fast-SCNN, a competitive (68.0\%) and above real-time semantic segmentation algorithm (123.5 fps) for high resolution images ($1024 \times 2048px$).
\item We adapt the skip connection, popular in offline DCNNs, and propose a shallow \textit{learning to downsample} module for fast and efficient multi-branch low-level feature extraction.
\item We specifically design Fast-SCNN to be of low capacity, and we empirically validate that running training for more epochs is equivalently successful  to pre-training with ImageNet or training with additional coarse data in our small capacity network.
\end{enumerate}
Moreover, we employ Fast-SCNN to subsampled input data, achieving state-of-the-art performance without the need for redesigning our network.

%-------------------------------------------------------------------------
\section{Related Work}
\label{sec:related-work}
We discuss and compare semantic image segmentation frameworks with a particular focus on real-time execution with low energy and memory requirements \cite{segnet-badrinarayanan2017,enet-paszke2016,contextnet-poudel2018,icnet-zhao2017b,BiSeNet-yu2018,gun-mazzini2018,erfnet-romera2018,espnet-mehta2018}.

\subsection{Foundation of Semantic Segmentation}

State-of-the-art semantic segmentation DCNNs combine two separate modules: the encoder and the decoder. The encoder module uses a combination of convolution and pooling operations to extract DCNN features. The decoder module recovers the spatial details from the sub-resolution features, and predicts the object labels (\ie the semantic segmentation) \cite{fcn-long2016,segnet-badrinarayanan2017}. Most commonly, the encoder is adapted from a simple classification DCNN method, such as VGG \cite{vgg-simonyan2014} or ResNet \cite{resnet-he2015}. In semantic segmentation, the fully connected layers are removed.

The seminal fully convolution network (FCN)~\cite{fcn-long2016} laid the foundation for most modern segmentation architectures. Specifically, FCN employs VGG \cite{vgg-simonyan2014} as encoder, and bilinear upsampling in combination with skip-connection from lower layers to recover spatial detail. U-Net \cite{u-net-ronneberger2015} further exploited the spatial details using dense skip connections.

Later, inspired by global image-level context prior to DCNNs \cite{lazebnik2006,lucchi2011}, the pyramid pooling module of PSPNet \cite{pspnet-zhao2017a} and atrous spatial pyramid pooling (ASPP) of DeepLab \cite{deeplab-v2-chen2016} are employed to encode and utilize global context.

Other competitive fundamental segmentation architectures use conditional random fields (CRF) \cite{zheng2015,chen2014} or recurrent neural networks \cite{visin2015,zheng2015}. However, none of them run in real-time.

Similar to the object detection \cite{yolo-redmon2016,yolo9000-redmon2016,ssd-liu2015}, speed became one important factor in image segmentation system design \cite{contextnet-poudel2018,BiSeNet-yu2018,gun-mazzini2018,erfnet-romera2018,icnet-zhao2017b,enet-paszke2016}. Building on FCN, SegNet \cite{segnet-badrinarayanan2017} introduced a joint encoder-decoder model and became one of the earliest efficient segmentation models. Following SegNet, ENet \cite{enet-paszke2016} also design an encoder-decoder with few layers to reduce the computational cost.

More recently, two-branch and multi-branch systems were introduced. ICNet \cite{icnet-zhao2017b}, ContextNet \cite{contextnet-poudel2018}, BiSeNet \cite{BiSeNet-yu2018} and GUN \cite{gun-mazzini2018} learned global context with reduced-resolution input in a deep branch, while boundaries are learned in a shallow branch at full resolution.

However, state-of-the-art real-time semantic segmentation remains challenging, and typically requires high-end GPUs. Inspired by two-branch methods, Fast-SCNN incorporates a shared shallow network path to encode detail, while context is efficiently learned at low resolution (Figure~\ref{fig:fast-scnn_compare}).

\subsection{Efficiency in DCNNs}

The common techniques of efficient DCNNs can be divided into four categories:

\textbf{Depthwise Separable Convolutions:} MobileNet~\cite{mobilenet-howard2017} decomposes a standard convolution into a depthwise convolution and a $1 \times 1$ pointwise convolution, together known as depthwise separable convolution. Such a factorization reduces the floating point operations and convolutional parameters, hence the computational cost and memory requirement of the model is reduced.

\textbf{Efficient Redesign of DCNNs:} Chollet \cite{xception-net-chollet2016} designed the Xception network using efficient depthwise separable convolution. MobleNet-V2 proposed inverted bottleneck residual blocks \cite{inverted-res-bottlenecks-sandler2018} to build an efficient DCNN for the classification task. ContextNet \cite{contextnet-poudel2018} used inverted bottleneck residual blocks to design a two-branch network for efficient real-time semantic segmentation. Similarly, \cite{BiSeNet-yu2018,gun-mazzini2018,icnet-zhao2017b} propose multi-branch segmentation networks to achieve real-time performance.
  
\textbf{Network Quantization:} Since floating point multiplications are costly compared to integer or binary operations, runtime can be further reduced using quantization techniques for DCNN filters and activation values \cite{hubara2016,xnornet-rastegari2016,wu2018}.
    
\textbf{Network Compression:} Pruning is applied to reduce the size of a pre-trained network, resulting in faster runtime, a smaller parameter set, and smaller memory footprint \cite{contextnet-poudel2018,deep-compression-han2016,pruning-fliters-li2017}.

\begin{figure}[t]
\begin{center}
   \includegraphics[width=1\linewidth]{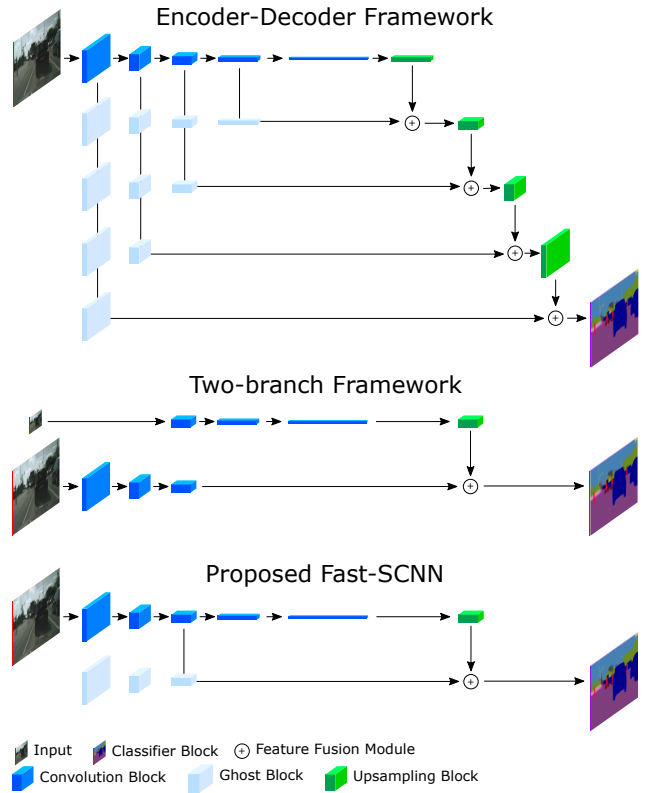}
\end{center}
   \caption{Schematic comparison of Fast-SCNN with encoder-decoder and two-branch architectures. Encoder-decoder employs multiple skip connections at many resolutions, often resulting from deep convolution blocks. Two-branch methods employ global features from low resolution with shallow spatial detail. Fast-SCNN encodes spatial detail and initial layers of global context in our learning to downsample module simultaneously.}
\label{fig:fast-scnn_compare}
\end{figure}

Fast-SCNN relies heavily on depthwise separable convolutions and residual bottleneck blocks \cite{inverted-res-bottlenecks-sandler2018}. Furthermore we introduce a two-branch model that incorporates our learning to downsample module, allowing for shared feature extraction at multiple resolution levels (Figure~\ref{fig:fast-scnn_compare}). Note, even though the initial layers of the  multiple branches extract similar features \cite{deconv-zeiler2014,olah2017}, common two-branch approaches do not leverage this. Network quantization and network compression can be applied orthogonally, and is left to future work.

\subsection{Pre-training on Auxiliary Tasks}
It is a common belief that pre-training on auxiliary tasks boosts system accuracy. Earlier works on object detection \cite{rcnn-girshick2013} and semantic segmentation \cite{deeplab-v2-chen2016,pspnet-zhao2017a} have shown this with pre-training on ImageNet \cite{imagenet2015}. Following this trend, other real-time efficient semantic segmentation methods are also pre-trained on ImageNet \cite{icnet-zhao2017b,BiSeNet-yu2018,gun-mazzini2018}. However, it is not known whether pre-training is necessary on low-capacity networks. Fast-SCNN is specifically designed with low capacity. In our experiments we show that small networks do not get significant benefit from pre-training. Instead, aggressive data augmentation and more number of epochs provide similar results.

%-------------------------------------------------------------------------
\section{Proposed Fast-SCNN}
\label{sec:fast-seg-net}

\begin{table}[t]
\begin{center}
\scalebox{1}{%
\begin{tabular}{l l l l l l}\hline
  Input & Block & t & c & n & s \\ \hline
  $1024\times2048\times3$  & Conv2D     & -  & 32  & 1 & 2 \\ %\hline
  $512\times1024\times32$  & DSConv     & -  & 48  & 1 & 2 \\ %\hline
  $256\times512\times48$  & DSConv     & -  & 64  & 1 & 2 \\ %\hline
  \hline
  $128\times256\times64$ & bottleneck & 6  & 64  & 3 & 2 \\ %\hline
  $64\times128\times64$ & bottleneck & 6  & 96  & 3 & 2 \\ %\hline
  $32\times64\times96$ & bottleneck & 6  & 128  & 3 & 1 \\ %\hline
  $32\times64\times128$   & PPM & -  & 128 & - & - \\ %\hline
  \hline
  $32\times64\times128$   & FFM & -  & 128 & - & - \\ %\hline
  \hline
  $128\times256\times128$   & DSConv  & -  & 128 & 2 & 1 \\ %\hline
  $128\times256\times128$   & Conv2D  & -  & 19 & 1 & 1 \\ %\hline
  \hline
\end{tabular}}
\end{center}
\caption{Fast-SCNN uses standard convolution (Conv2D), depthwise separable convolution (DSConv), inverted residual bottleneck blocks (bottleneck), a pyramid pooling module (PPM) and a feature fusion module (FFM) block. Parameters t, c, n and s represent expansion factor of the bottleneck block, number of output channels, number of times block is repeated and stride parameter which is applied to first sequence of the repeating block. The horizontal lines separate the modules: learning to down-sample, global feature extractor, feature fusion and classifier (top to bottom).}
\label{tbl:fast-scnn}
\end{table}
Fast-SCNN is inspired by the two-branch architectures \cite{contextnet-poudel2018,BiSeNet-yu2018,gun-mazzini2018} and encoder-decoder networks with skip connections \cite{fcn-long2016,u-net-ronneberger2015}. Noting that early layers commonly extract low-level features. We reinterpret skip connections as a learning to downsample module, enabling us to merge the key ideas of both frameworks, and allowing us to build a fast semantic segmentation model. Figure \ref{fig:fast-scnn} and Table~\ref{tbl:fast-scnn} present the layout of Fast-SCNN. In the following we discuss our motivation and describe our building blocks in more detail.

\subsection{Motivation}
Current state-of-the-art semantic segmentation methods that run in real-time are based on networks with two branches, each operating on a different resolution level \cite{contextnet-poudel2018,BiSeNet-yu2018,gun-mazzini2018}. They learn global information from low-resolution versions of the input image, and shallow networks at full input resolution are employed to refine the precision of the segmentation results. Since input resolution and network depth are main factors for runtime, these two-branch approaches allow for real-time computation.

It is well known that the first few layers of DCNNs extract the low-level features, such as edges and corners \cite{deconv-zeiler2014,olah2017}. Therefore, rather than employing a two-branch approach with separate computation, we introduce learning to downsample, which shares feature computation between the low and high-level branch in a shallow network block.

\subsection{Network Architecture}
Our Fast-SCNN uses a learning to downsample module, a coarse global feature extractor, a feature fusion module and a standard classifier. All modules are built using depthwise separable convolution, which has become a key building block for many efficient DCNN architectures \cite{xception-net-chollet2016,mobilenet-howard2017,contextnet-poudel2018}.

\subsubsection{Learning to Downsample}
In our learning to downsample module, we employ three layers. Only three layers are employed to ensure low-level feature sharing is valid, and efficiently implemented. The first layer is a standard convolutional layer (Conv2D) and the remaining two layers are depthwise separable convolutional layers (DSConv). Here we emphasize, although DSConv is computationally more efficient, we employ Conv2D since the input image only has three channels, making DSConv's computational benefit insignificant at this stage.

All three layers in our learning to downsample module use stride 2, followed by batch normalization \cite{batch-norm-ioffe2015} and ReLU. The spatial kernel size of the convolutional and depthwise layers is $3\times3$. Following \cite{xception-net-chollet2016,inverted-res-bottlenecks-sandler2018,contextnet-poudel2018}, we omit the nonlinearity between depthwise and pointwise convolutions.

\subsubsection{Global Feature Extractor}
\begin{table}[t]
\begin{center}
\scalebox{1}{%
        \begin{tabular}{c c c} \hline
        Input & Operator& Output \\ \hline
        $h \times w \times c$  & Conv2D $1/1,f$  & $h \times w \times tc$ \\
        $h \times w \times tc$ & DWConv $3/s,f$ & $\frac{h}{s} \times \frac{w}{s} \times tc$ \\
        $\frac{h}{s} \times \frac{w}{s} \times tc$ & Conv2D $1/1,-$ & $\frac{h}{s} \times \frac{w}{s} \times {c}'$ \\
        \hline
\end{tabular}}
\end{center}
\caption{The \textit{bottleneck residual block} transfers the input from $c$ to ${c}'$ channels with expansion factor $t$. Note, the last pointwise convolution does not use non-linearity $f$. The input is of height $h$ and width $w$, and x/$s$ represents kernel size and stride of the layer.}
\label{tbl:inverted-residual-block}
\end{table}
The global feature extractor module is aimed at capturing the global context for image segmentation. In contrast to common two-branch methods which operate on low-resolution versions of the input image, our module directly takes the output of the learning to downsample module (which is at $\frac{1}{8}$-resolution of the original input). The detailed structure of the module is shown in Table~\ref{tbl:fast-scnn}. We use efficient bottleneck residual block introduced by MobileNet-V2 \cite{inverted-res-bottlenecks-sandler2018} (Table \ref{tbl:inverted-residual-block}). In particular, we employ residual connection for the bottleneck residual blocks when the input and output are of the same size. Our bottleneck block uses an efficient depthwise separable convolution, resulting in less number of parameters and floating point operations. Also, a pyramid pooling module (PPM) \cite{pspnet-zhao2017a} is added at the end to aggregate the different-region-based context information.

\subsubsection{Feature Fusion Module}
\begin{table}[t]
\begin{center}
\scalebox{1}{%
        \begin{tabular}{c c} \hline
        Higher resolution & X times lower resolution \\ \hline
        \hspace{4em}-\hspace{4em}~& Upsample $\times$ X \\
        - & DWConv (dilation X) $3/1,f$ \\
        Conv2D $1/1,-$ &  Conv2D $1/1,-$ \\ \hline
        \multicolumn{2}{c}{add,$f$} \\
        \hline
\end{tabular}}
\end{center}
\caption{Features fusion module (FFM) of Fast-SCNN. Note, the pointwise convolutions are of desired output, and do not use non-linearity $f$. Non-linearity $f$ is employed after adding the features.}
\label{tbl:feature-fusion-module}
\end{table}
Similar to ICNet \cite{icnet-zhao2017b} and ContextNet \cite{contextnet-poudel2018} we prefer simple addition of the features to ensure efficiency. Alternatively, more sophisticated feature fusion modules (\eg \cite{BiSeNet-yu2018}) could be employed at the cost of runtime performance, to reach better accuracy. The detail of the feature fusion module is shown in Table~\ref{tbl:feature-fusion-module}.

\subsubsection{Classifier}
In the classifier we employ two depthwise separable convolutions (DSConv) and one pointwise convolution (Conv2D). We found that adding few layers after the feature fusion module boosts the accuracy. The details of the classifier module is shown in the Table \ref{tbl:fast-scnn}. 

Softmax is used during training, since gradient decent is employed. During inference we may substitute costly softmax computations with argmax, since both functions are monotonically increasing. We denote this option as Fast-SCNN cls (classification). On the other hand, if a standard DCNN based probabilistic model is desired, softmax is used, denoted as Fast-SCNN prob (probability).

\subsection{Comparison with Prior Art}
Our model is inspired by the two-branch framework, and incorporates ideas of encoder-decorder methods (Figure~\ref{fig:fast-scnn_compare}).

\subsubsection{Relation with Two-branch Models}
The state-of-the-art real-time models (ContextNet \cite{contextnet-poudel2018}, BiSeNet \cite{BiSeNet-yu2018} and GUN \cite{gun-mazzini2018}) use two-branch networks. Our learning to downsample module is equivalent to their spatial path, as it is shallow, learns from full resolution, and is used in the feature fusion module (Figure~\ref{fig:fast-scnn}).

Our global feature extractor module is equivalent to the deeper low-resolution branch of such approaches. In contrast, our global feature extractor shares its computation of the first few layers with the learning to downsample module. By sharing the layers we not only reduce computational complexity of feature extraction, but we also reduce the required input size as Fast-SCNN uses $\frac{1}{8}$-resolution instead of $\frac{1}{4}$-resolution for global feature extraction.

\subsubsection{Relation with Encoder-Decoder Models}
Proposed Fast-SCNN can be viewed as a special case of an encoder-decoder framework, such as FCN \cite{fcn-long2016} or U-Net \cite{u-net-ronneberger2015}. However, unlike the multiple skip connections in FCN and the dense skip connections in U-Net, Fast-SCNN only employs a single skip connection to reduce computations as well as memory.

In correspondence with \cite{deconv-zeiler2014}, who advocate that features are shared only at early layers in DCNNs, we position our skip connection early in our network. In contrast, prior art typically employ deeper modules at each resolution, before skip connections are applied.

%-------------------------------------------------------------------------
\section{Experiments}
\label{sec:experiments}

We evaluated our proposed fast segmentation convolutional neural network (Fast-SCNN) on the validation set of the Cityscapes dataset \cite{cityscaples2016}, and report its performance on the Cityscapes test set, \ie the Cityscapes benchmark server.

\subsection{Implementation Details}
Implementation detail is as important as theory when it comes to efficient DCNNs. Hence, we carefully describe our setup here. We conduct experiments on the TensorFlow machine learning platform using Python. Our experiments are executed on a workstation with either Nvidia Titan X (Maxwell) or Nvidia Titan Xp (Pascal) GPU, with CUDA 9.0 and CuDNN v7. Runtime evaluation is performed in a single CPU thread and one GPU to measure the forward inference time. We use 100 frames for burn-in and report average of 100 frames for the frames per second (fps) measurement.

We use stochastic gradient decent (SGD) with momentum 0.9 and batch-size 12. Inspired by \cite{deeplab-v2-chen2016,pspnet-zhao2017a,mobilenet-howard2017} we use `poly' learning rate with the base one as 0.045 and power as 0.9. Similar to MobileNet-V2 we found that $\ell_2$ regularization is not necessary on depthwise convolutions, for other layers $\ell_2$ is 0.00004. Since training data for semantic segmentation is limited, we apply various data augmentation techniques: random resizing between 0.5 to 2, translation/crop, horizontal flip, color channels noise and brightness. Our model is trained with cross-entropy loss. We found that auxiliary losses at the end of learning to downsample and the global feature extraction modules with 0.4 weights are beneficial.

Batch normalization \cite{batch-norm-ioffe2015} is used before every non-linear function. Dropout is used only on the last layer, just before the softmax layer. Contrary to MobileNet \cite{mobilenet-howard2017} and ContextNet \cite{contextnet-poudel2018}, we found that Fast-SCNN trains faster with ReLU and achieves slightly better accuracy than ReLU6, even with the depthwise separable convolutions that we use throughout our model.

We found that the performance of DCNNs can be improved by training for higher number of iterations, hence we train our model for 1,000 epochs unless otherwise stated, using the Cityescapes dataset \cite{cityscaples2016}. It is worth noting here, Fast-SCNN's capacity is deliberately very low, as we employ 1.11 million parameters. Later we show that aggressive data augmentation techniques make overfitting unlikely.

\subsection{Evaluation on Cityscapes}
\begin{table}[tb]
\begin{center}
\scalebox{0.9}{%
\begin{tabular}{l c c c c c}
\hline
\textbf{Model} & \textbf{Class} & \textbf{Category} & \textbf{Params} \\
\hline
	DeepLab-v2 \cite{deeplab-v2-chen2016}* & 70.4 & 86.4 & 44.-- \\ %\hline
    PSPNet \cite{pspnet-zhao2017a}* & 78.4 & 90.6 & 65.7\_ \\ %\hline
    \hline
    SegNet \cite{segnet-badrinarayanan2017} & 56.1 & 79.8 & 29.46 \\ %\hline
    ENet \cite{enet-paszke2016} & 58.3 & 80.4 & 00.37 \\ %\hline
    ICNet \cite{icnet-zhao2017b}* & 69.5 & - & 06.68 \\ %\hline
    ERFNet \cite{erfnet-romera2018} & 68.0 & 86.5 & 02.1\_ \\ %\hline
    %ESPNet \cite{mehta2018} & 60.3 & 82.2 & 0.4X \\ \hline
    BiSeNet \cite{BiSeNet-yu2018} & 71.4 & - & 05.8\_ \\ %\hline
    GUN \cite{gun-mazzini2018} & 70.4 & - & - \\ %\hline
    ContextNet \cite{contextnet-poudel2018} & 66.1 & 82.7 & 00.85 \\ %\hline
    Fast-SCNN (Ours) & 68.0 & 84.7 & 01.11 \\ %\hline
\hline
\end{tabular}}
\end{center}
\caption{Class and category mIoU of the proposed Fast-SCNN compared to other state-of-the-art semantic segmentation methods on the Cityscapes test set. Number of parameters is listed in millions.}
\label{tbl:comparision-cityscapes-test-set}
\end{table}
\begin{table}[tb]
\begin{center}
\scalebox{0.9}{%
\begin{tabular}{l c c c}
\hline
    ~ & $1024\times2048$  & $512\times1024$ & $256\times512$ \\ \hline
    SegNet \cite{segnet-badrinarayanan2017} & 1.6 & - & - \\ %\hline
    ENet \cite{enet-paszke2016} & 20.4 & 76.9 & 142.9 \\ %\hline
    ICNet \cite{icnet-zhao2017b} & 30.3 & - & - \\ %\hline
    ERFNet \cite{erfnet-romera2018} & 11.2 & 41.7 & 125.0\\ %\hline
    %ESPNet \cite{mehta2018} & x.xx & x.xx \\ \hline
    %ContextNet \cite{poudel2018} & 18.3 & 65.5 & 139.2 \\ \hline
    ContextNet \cite{contextnet-poudel2018} & 41.9 & 136.2 & 299.5 \\ %\hline
    Our prob  & 62.1 & 197.6 & 372.8 \\ %\hline
    Our cls  & 75.3 & 218.8 & 403.1 \\ %\hline
    \hline
    BiSeNet \cite{BiSeNet-yu2018}* & 57.3 & - & - \\ %\hline
    GUN \cite{gun-mazzini2018}* & 33.3 & - & - \\ %\hline
    % Titan 1080 Ti
    %Our prob*  & 67.4 & 209.4 & 456.8 \\
    %Our cls*  & 77.9 & 231.2 & 483.2 \\
    % Titan X (Pascal)
    %Our prob*  & 69.6 & 213.0 & 465.6 \\
    %Our cls*  & 79.8 & 240.1 & 515.9 \\
    % Titan Xp (Pascal)
    Our prob*  & 106.2 & 266.3 & 432.9 \\
    Our cls*  & 123.5 & 285.8 & 485.4 \\
\hline
\end{tabular}}
\end{center}
\caption{Runtime (fps) on Nvidia Titan X (Maxwell, 3,072 CUDA cores) with TensorFlow~\cite{tensorflow2015}. Methods with `*' represent results on Nvidia Titan Xp (Pascal, 3,840 CUDA cores). Two versions of Fast-SCNN are shown: softmax output (our prob), and object label output (our cls).}
  \label{tbl:comparision-cityscapes-time}
\end{table}
\begin{figure*}[t]
\begin{center}
   \includegraphics[width=0.32\linewidth]{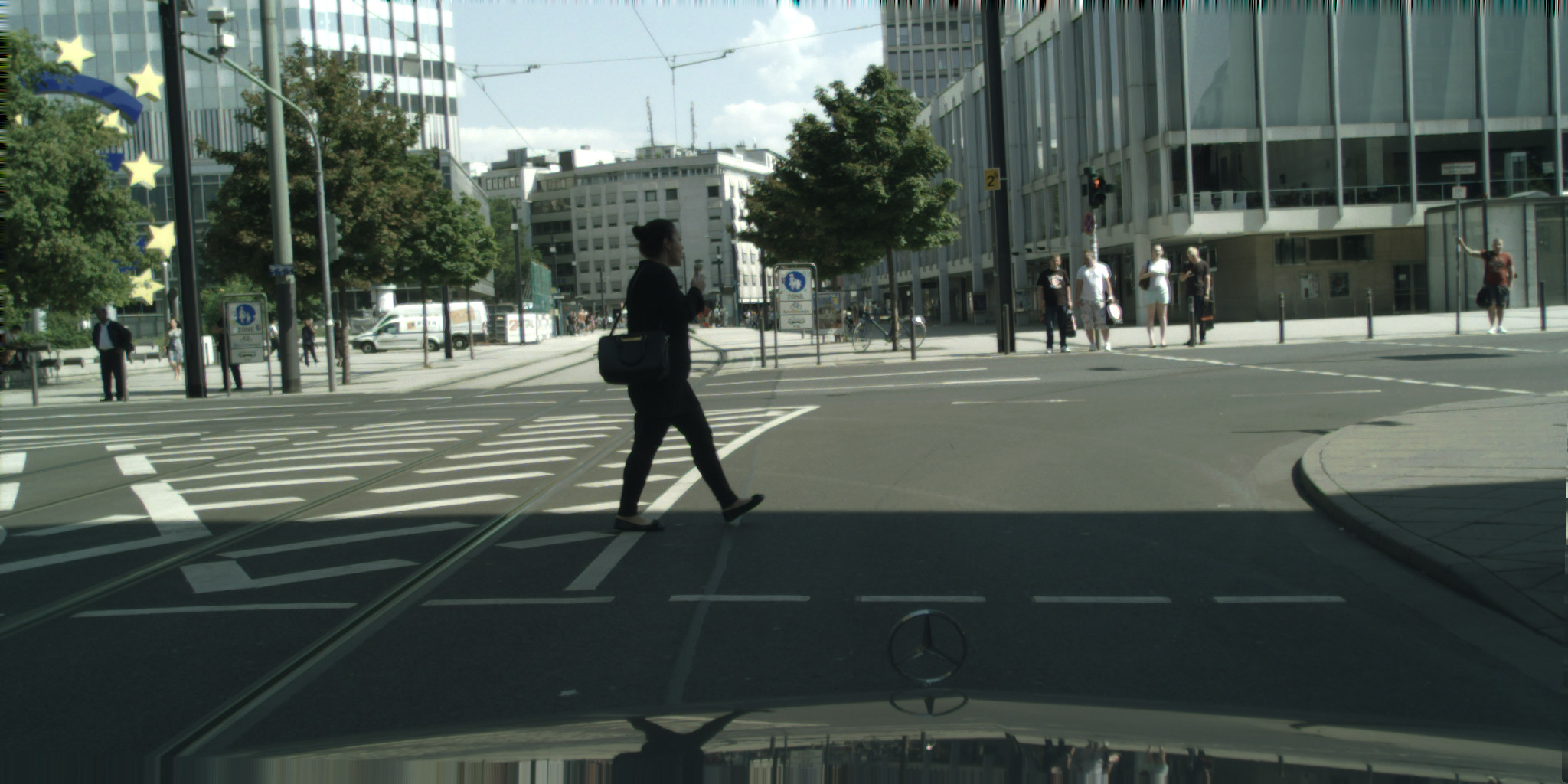}
   \includegraphics[width=0.32\linewidth]{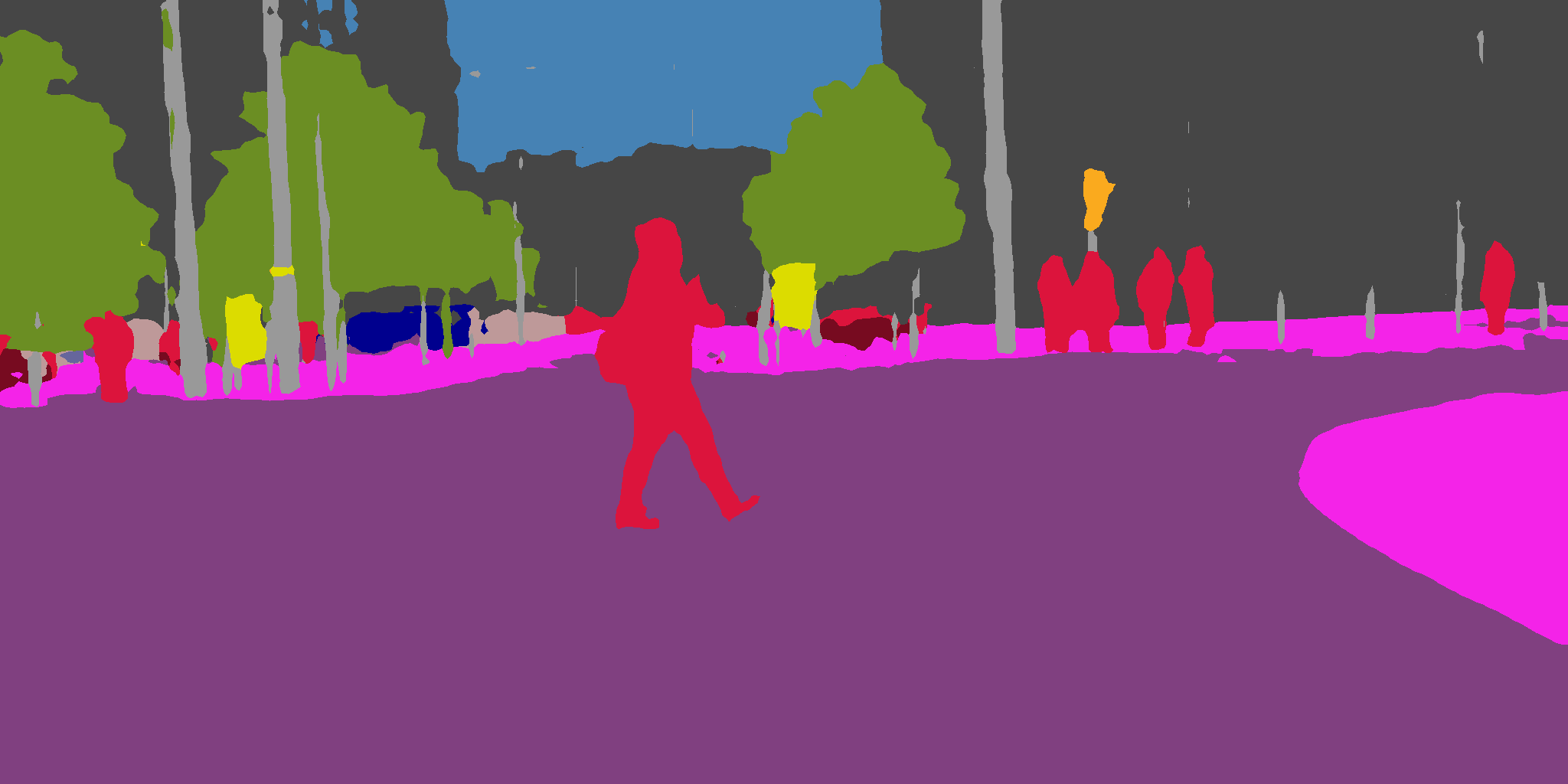}
   \includegraphics[width=0.32\linewidth]{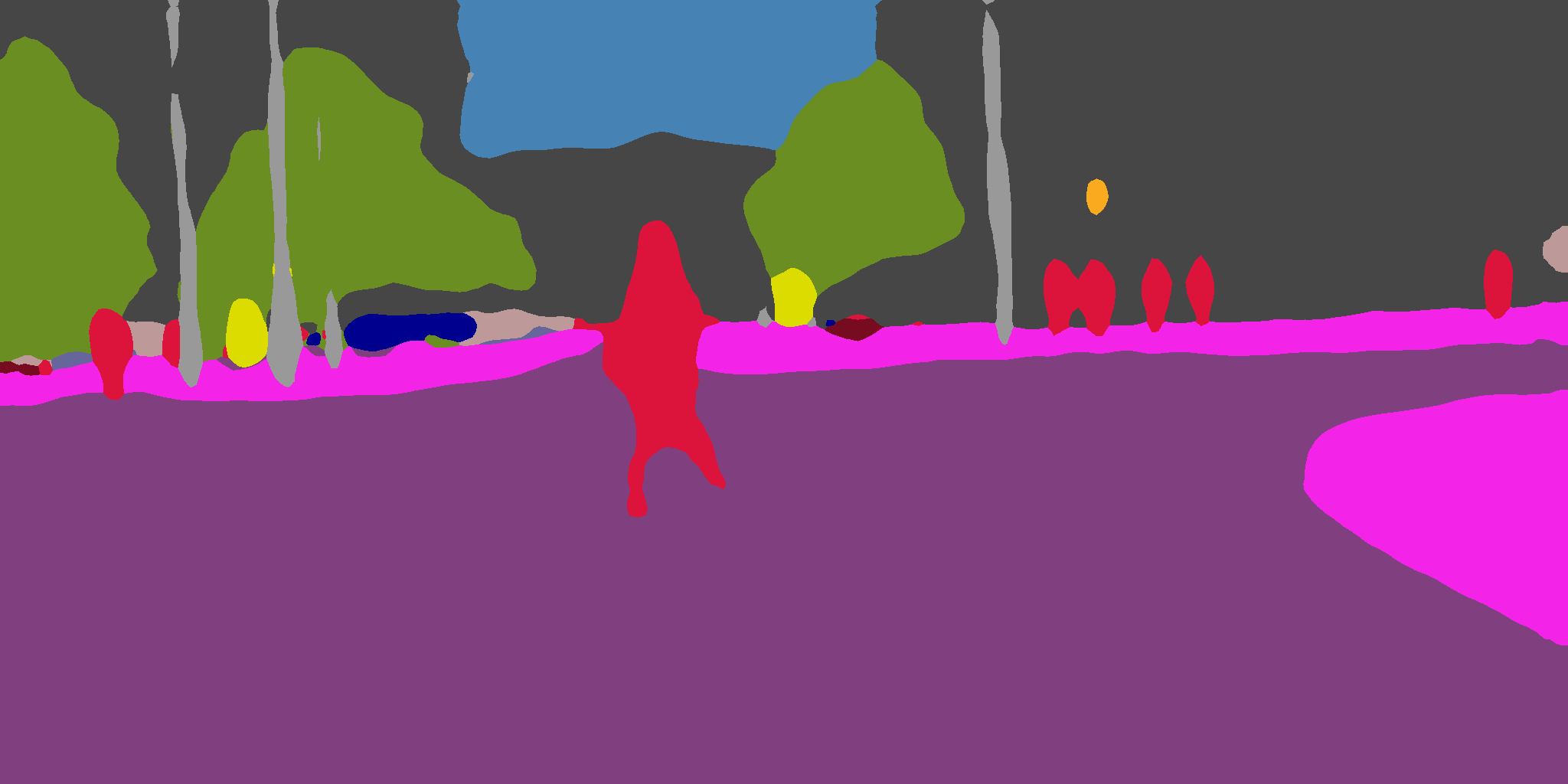}
   \\
    \includegraphics[width=0.32\linewidth]{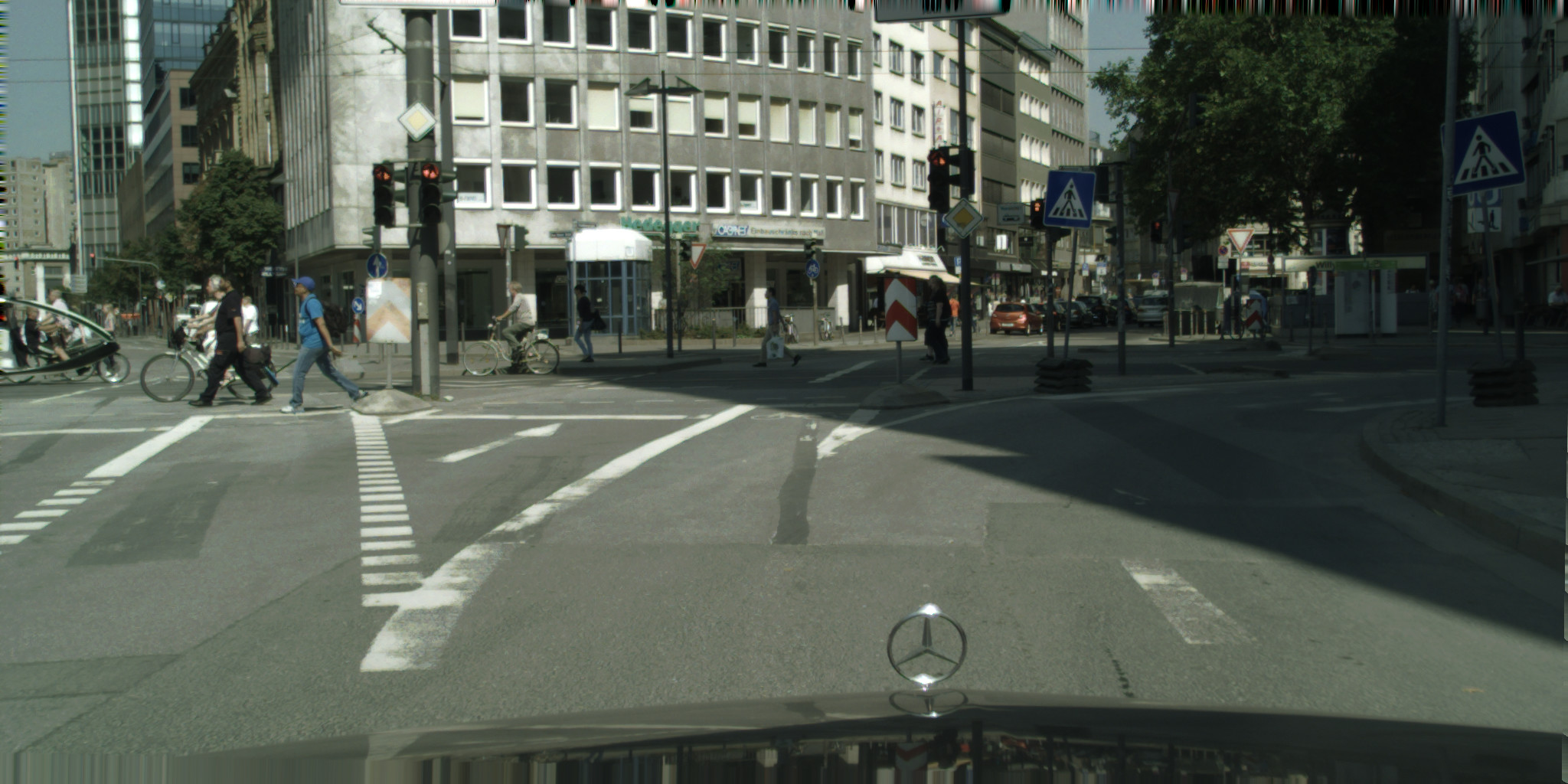}
   \includegraphics[width=0.32\linewidth]{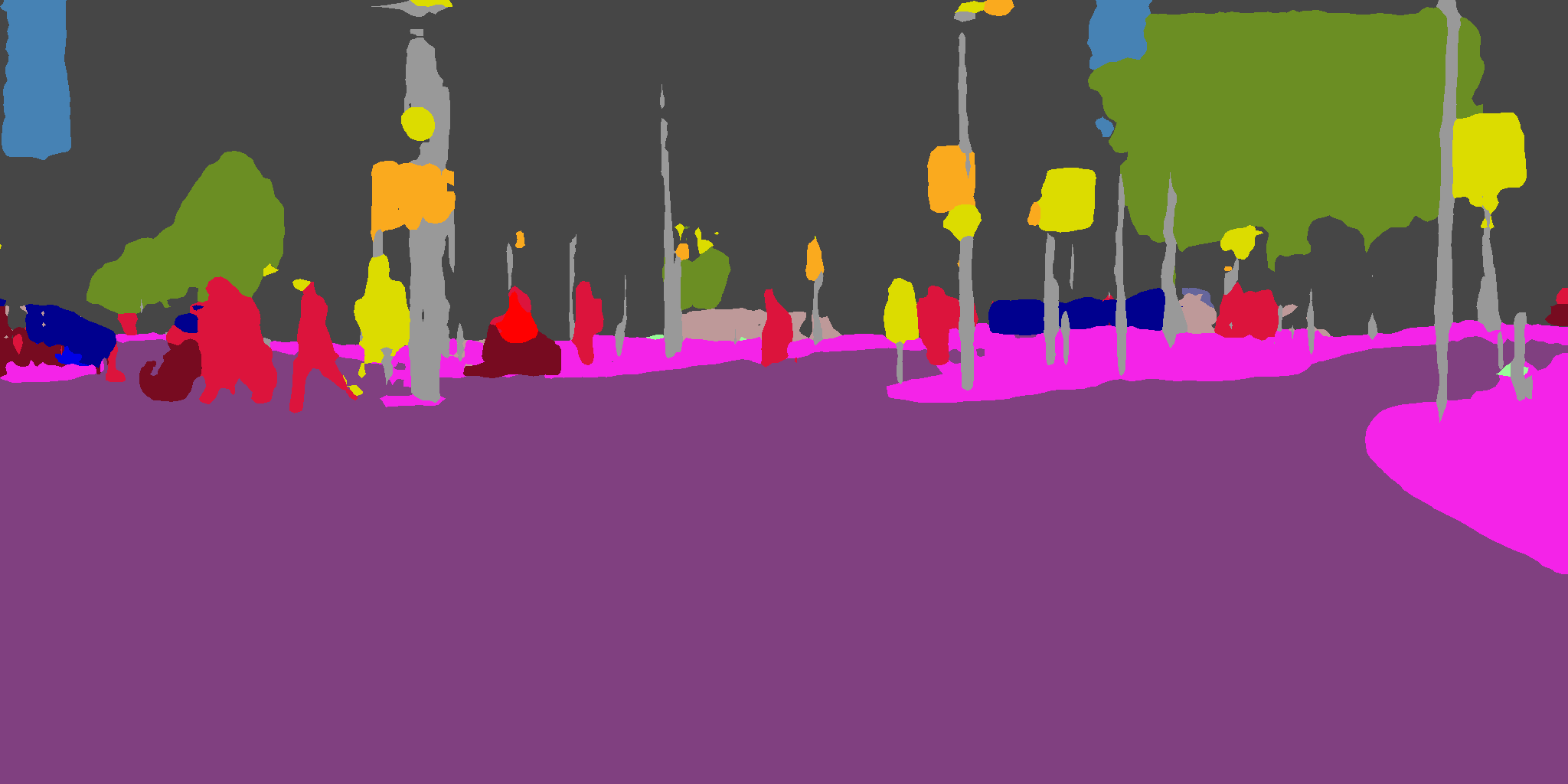}
   \includegraphics[width=0.32\linewidth]{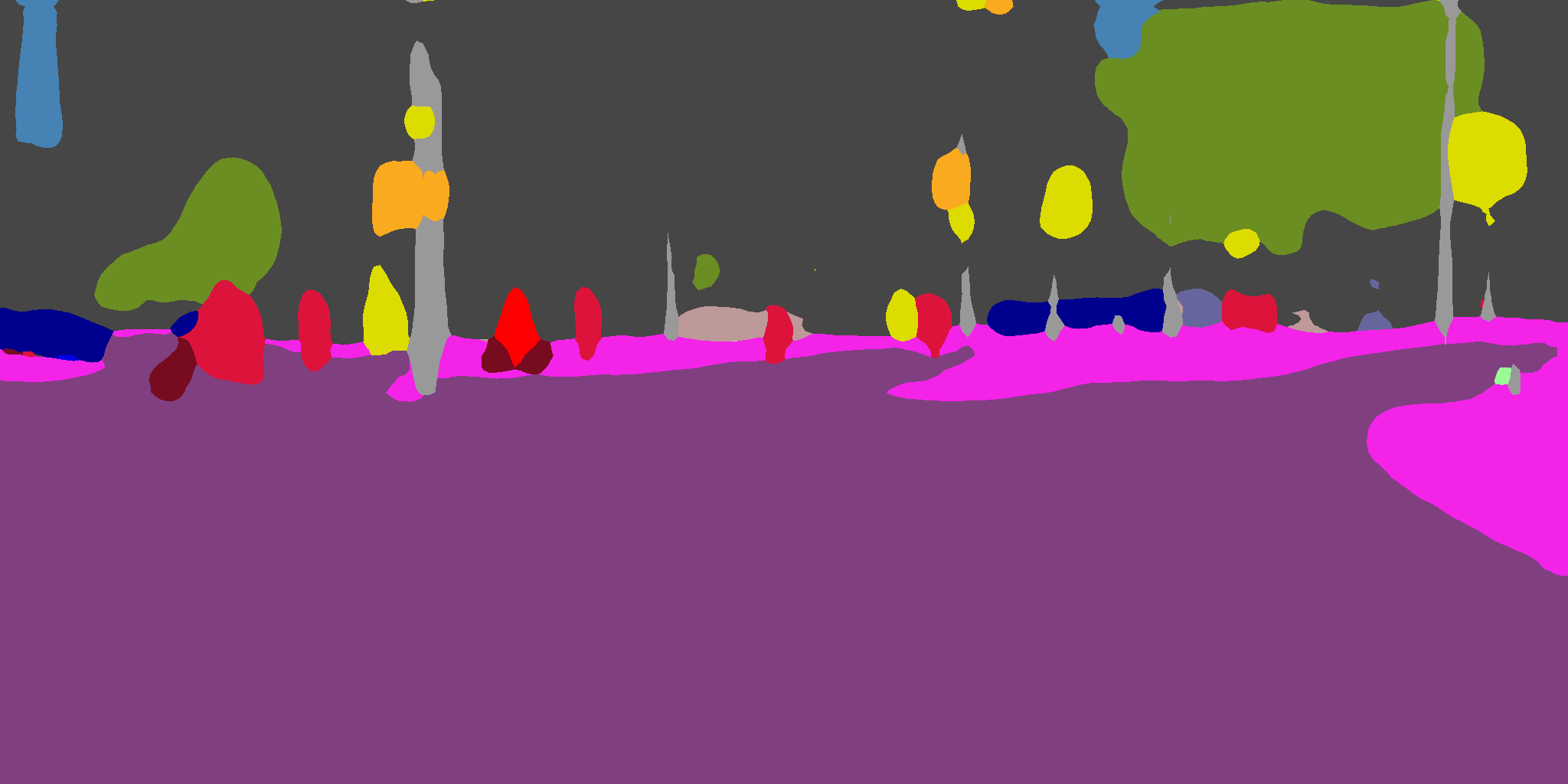}
   \\
    \includegraphics[width=0.32\linewidth]{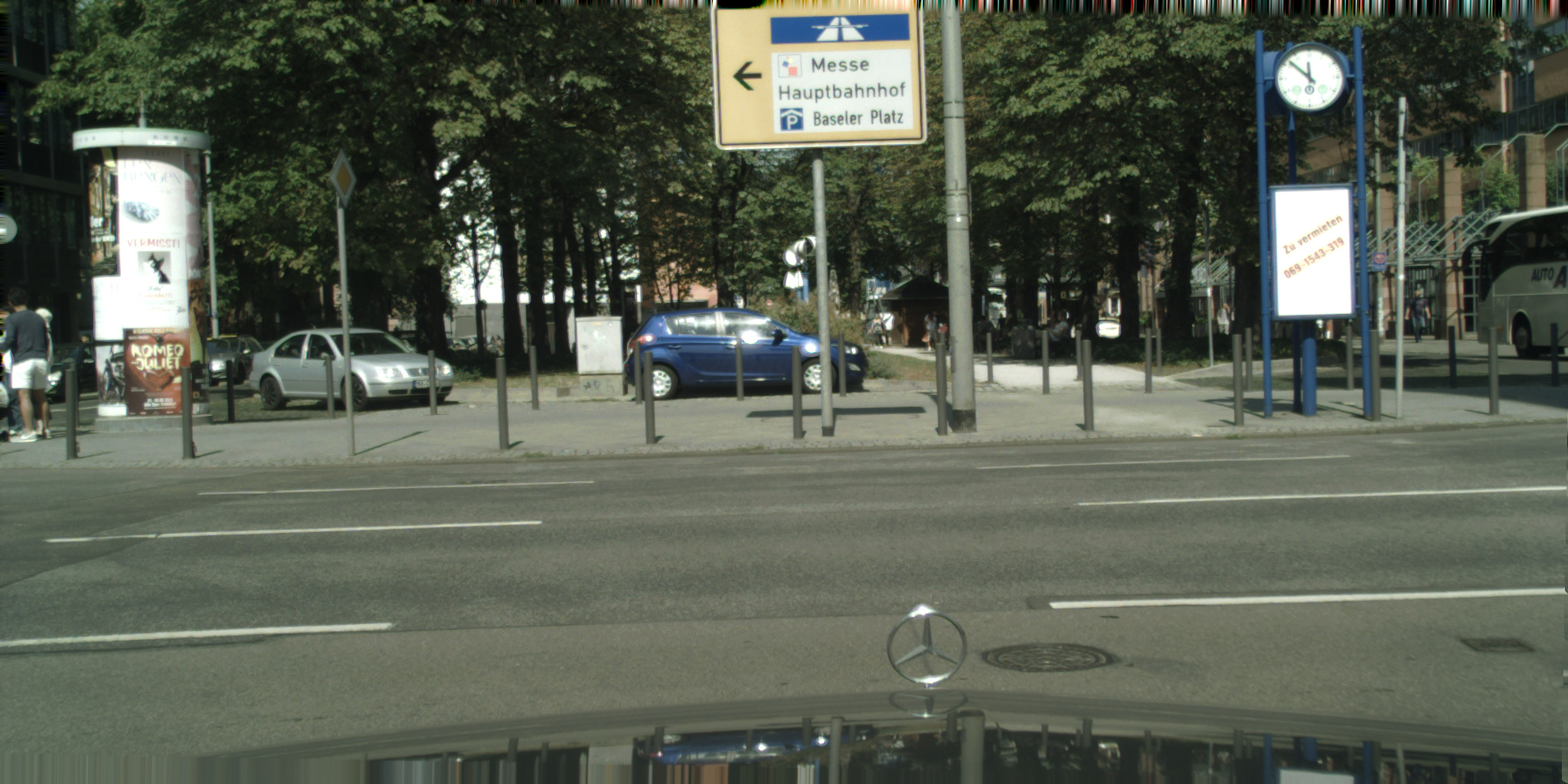}
   \includegraphics[width=0.32\linewidth]{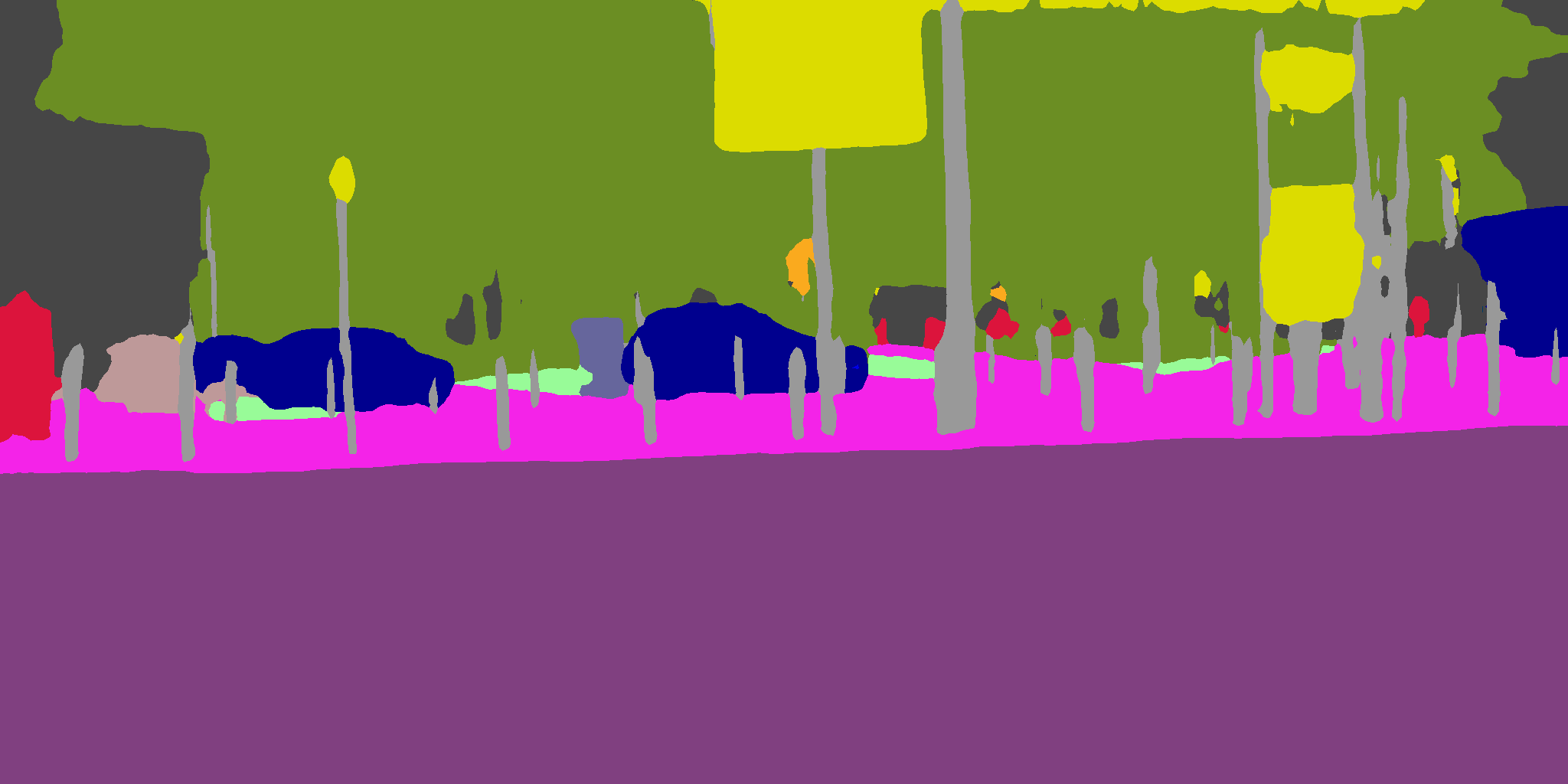}
   \includegraphics[width=0.32\linewidth]{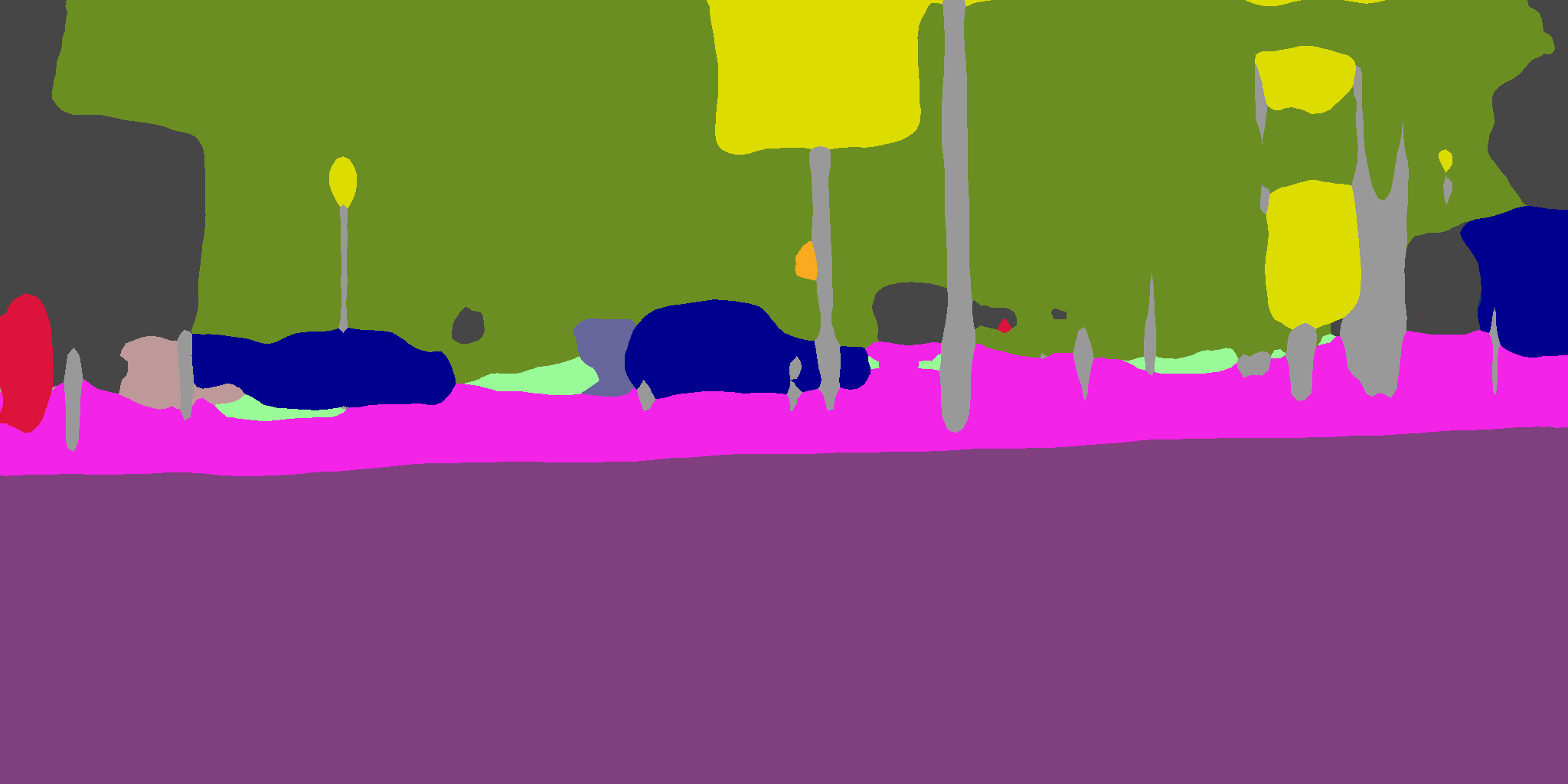}
\end{center}
   \caption{Visualization of Fast-SCNN's segmentation results. First column: input RGB images; second column: outputs of Fast-SCNN; and last column: outputs of Fast-SCNN after zeroing-out the contribution of the skip connection. In all results, Fast-SCNN benefits from skip connections especially at boundaries and objects of small size.}
\label{fig:deatil-branch-contribution}
\end{figure*}
We evaluate our proposed Fast-SCNN on Cityscapes, the largest publicly available dataset on urban roads \cite{cityscaples2016}. This dataset contains a diverse set of high resolution images ($1024 \times 2048px$) captured from 50 different cities in Europe. It has 5,000 images with high label quality: a training set of 2,975, validation set of 500 and test set of 1,525 images. The label for the training set and validation set are available and test results can be evaluated on the evaluation server. Additionally, 20,000 weakly annotated images (coarse labels) are available for training. We report results with both, fine only and fine with coarse labeled data. Cityscapes provides 30 class labels, while only 19 classes are used for evaluation. The mean of intersection over union (mIoU), and network inference time are reported in the following.

We evaluate overall performance on the withheld test set of Cityscapes \cite{cityscaples2016}. The comparison between the proposed Fast-SCNN and other state-of-the-art real-time semantic segmentation methods (ContextNet \cite{contextnet-poudel2018}, BiSeNet \cite{BiSeNet-yu2018}, GUN \cite{gun-mazzini2018}, ENet \cite{enet-paszke2016} and ICNet \cite{icnet-zhao2017b}) and offline methods (PSPNet \cite{pspnet-zhao2017a} and DeepLab-V2 \cite{deeplab-v2-chen2016}) is shown in Table~\ref{tbl:comparision-cityscapes-test-set}. Fast-SCNN achieves 68.0\% mIoU, which is slightly lower than BiSeNet (71.5\%) and GUN (70.4\%). ContextNet only achieves 66.1\% here.

Table~\ref{tbl:comparision-cityscapes-time} compares runtime at different resolutions. Here, BiSeNet (57.3 fps) and GUN (33.3 fps) are significantly slower than Fast-SCNN (123.5 fps). Compared to ContextNet (41.9 fps), Fast-SCNN is also significantly faster on Nvidia Titan X (Maxwell). Therefore we conclude, Fast-SCNN significantly improves upon state-of-the-art runtime with minor loss in accuracy. At this point we emphasize, our model is designed for low memory embedded devices. Fast-SCNN uses 1.11 million parameters, that is five times less than the competing BiSeNet at 5.8 million.

Finally, we zero-out the contribution of the skip connection and measure Fast-SCNN's performance. The mIoU reduced from 69.22\% to 64.30\% on the validation set. The qualitative results are compared in Figure~\ref{fig:deatil-branch-contribution}. As expected, Fast-SCNN benefits from the skip connection, especially around boundaries and objects of small size.

\subsection{Pre-training and Weakly Labeled Data}
\begin{table}[tb]
\begin{center}
\begin{tabular}{l c}
\hline
\textbf{Model} & \textbf{Class} \\
\hline
	ContextNet \cite{contextnet-poudel2018} & 65.9\_ \\ %\hline
	Fast-SCNN & 68.62 \\ %\hline
	Fast-SCNN + ImageNet & 69.15 \\ %\hline
	Fast-SCNN + Coarse & 69.22 \\ %\hline
	Fast-SCNN + Coarse + ImageNet & 69.19 \\ \hline
	%\hline
	%Fast-SCNN + 2 $\times$ num filters & 69.89 \\ \hline
\end{tabular}
\end{center}
\caption{Class mIoU of different Fast-SCNN settings on the Cityscapes validation set.}
\label{tbl:comparision-cityscapes-val-set}
\end{table}
High capacity DCNNs, such as R-CNN \cite{rcnn-girshick2013} and PSPNet \cite{pspnet-zhao2017a}, have shown that performance can be boosted with pre-training through different auxiliary tasks.
%Inspired by recent findings in \cite{rethinking-pruning_liu2018}, which suggest that low capacity networks are less likely to overfit and thus generalize better for pruning techniques,
As we specifically design Fast-SCNN to have low capacity, we now want to test performance with and without pre-training, and in connection with and without additional weakly labeled data. To the best of our knowledge, the significance of pre-training and additional weakly labeled data on low capacity DCNNs has not been studied before. Table~\ref{tbl:comparision-cityscapes-val-set} shows the results.

We pre-train Fast-SCNN on ImageNet \cite{imagenet2015} by replacing the feature fusion module with average pooling and the classification module now has a softmax layer only. Fast-SCNN achieves 60.71\% top-1 and 83.0\% top-5 accuracies on the ImageNet validation set. This result indicates that Fast-SCNN has insufficient capacity to reach comparable performance to most standard DCNNs on ImageNet ($>$70\% top-1) \cite{mobilenet-howard2017,inverted-res-bottlenecks-sandler2018}. The accuracy of Fast-SCNN with ImageNet pre-training yields 69.15\% mIoU on the validation set of Cityscapes, only 0.53\% improvement over Fast-SCNN without pre-training. Therefore we conclude, no significant boost can be achieved with ImageNet pre-training in Fast-SCNN.

Since the overlap between Cityscapes' urban roads and ImageNet's classification task is limited, it is reasonable to assume that Fast-SCNN may not benefit due to limited capacity for both domains. Therefore, we now incorporate the 20,000 coarsely labeled additional images provided by Cityscapes, as these are from a similar domain. Nevertheless, Fast-SCNN trained with coarse training data (with or without ImageNet) perform similar to each other, and only slightly improve upon the original Fast-SCNN without pre-training. Please note, small variations are insignificant and due to random initializations of the DCNNs.

It is worth noting here that working with auxiliary tasks is non-trivial as it requires architectural modifications in the network. Furthermore, licence restrictions and lack of resources further limit such setups. These costs can be saved, since we show that neither ImageNet pre-training nor weakly labeled data are significantly beneficial for our low capacity DCNN.
\begin{figure}[t]
\includegraphics[width=0.97\linewidth]{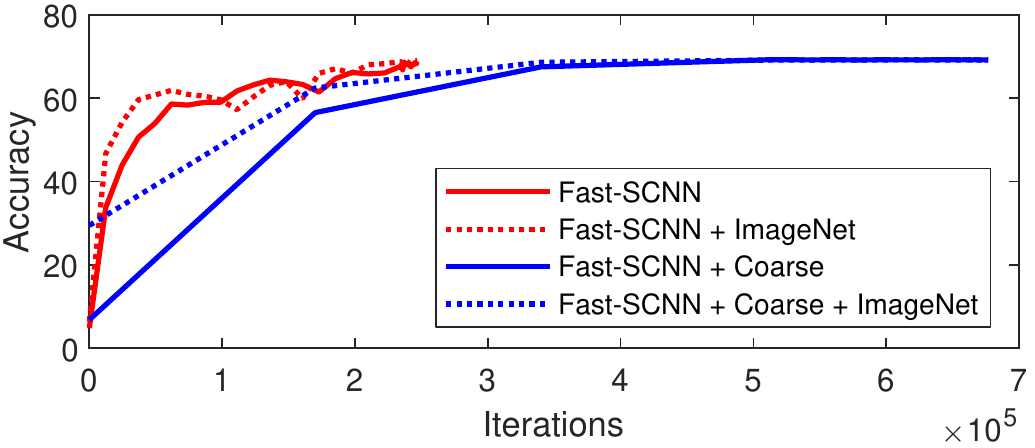}\\
\includegraphics[width=0.99\linewidth]{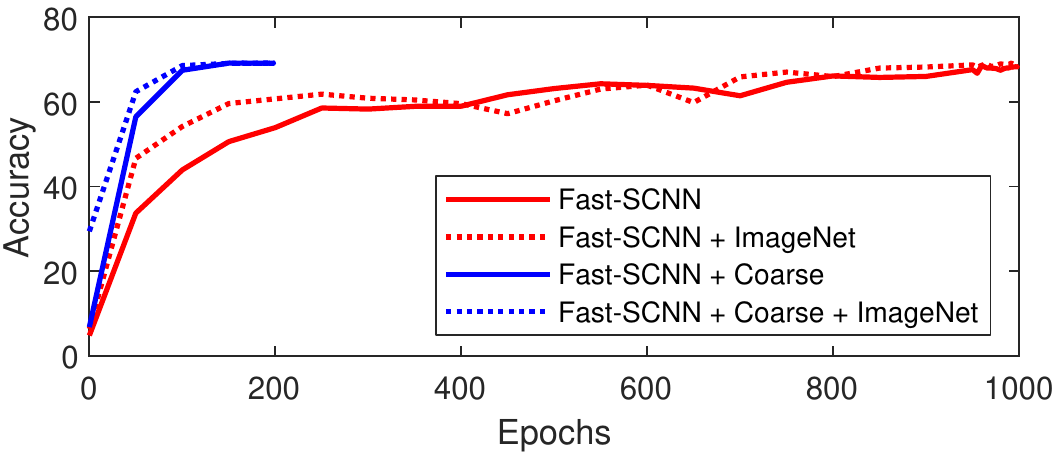}
\caption{Training curves on Cityscapes. Accuracy over iterations (top), and accuracy over epochs are shown (bottom). Dash lines represent ImageNet pre-training of the Fast-SCNN.}
\label{fig:iterations-vs-class-miou}
\end{figure}
Figure \ref{fig:iterations-vs-class-miou} shows the training curves. Fast-SCNN with coarse data trains slow in terms of iterations because of the weak label quality. Both ImageNet pre-trained versions perform better for early epochs (upto 400 epochs for training set alone, and 100 epochs when trained with the additional coarse labeled data). This means, we only need to train our model for longer to reach similar accuracy when we train our model from scratch.%, confirming similar finding by \cite{rethinking-pruning_liu2018}.
%Hence, this leads us to conclude that if trained longer enough it is possible to achieve accuracy similar to the large-scale pre-train version of the model. However, even though Fast-SCNN performs better with longer 1,000 epochs of the training, how can we ensure that it generalizes well with domain shit (test-set vs real data autonomous system come across) is left as a future work.
\begin{figure*}[t]
\begin{center}
   \includegraphics[width=0.32\linewidth]{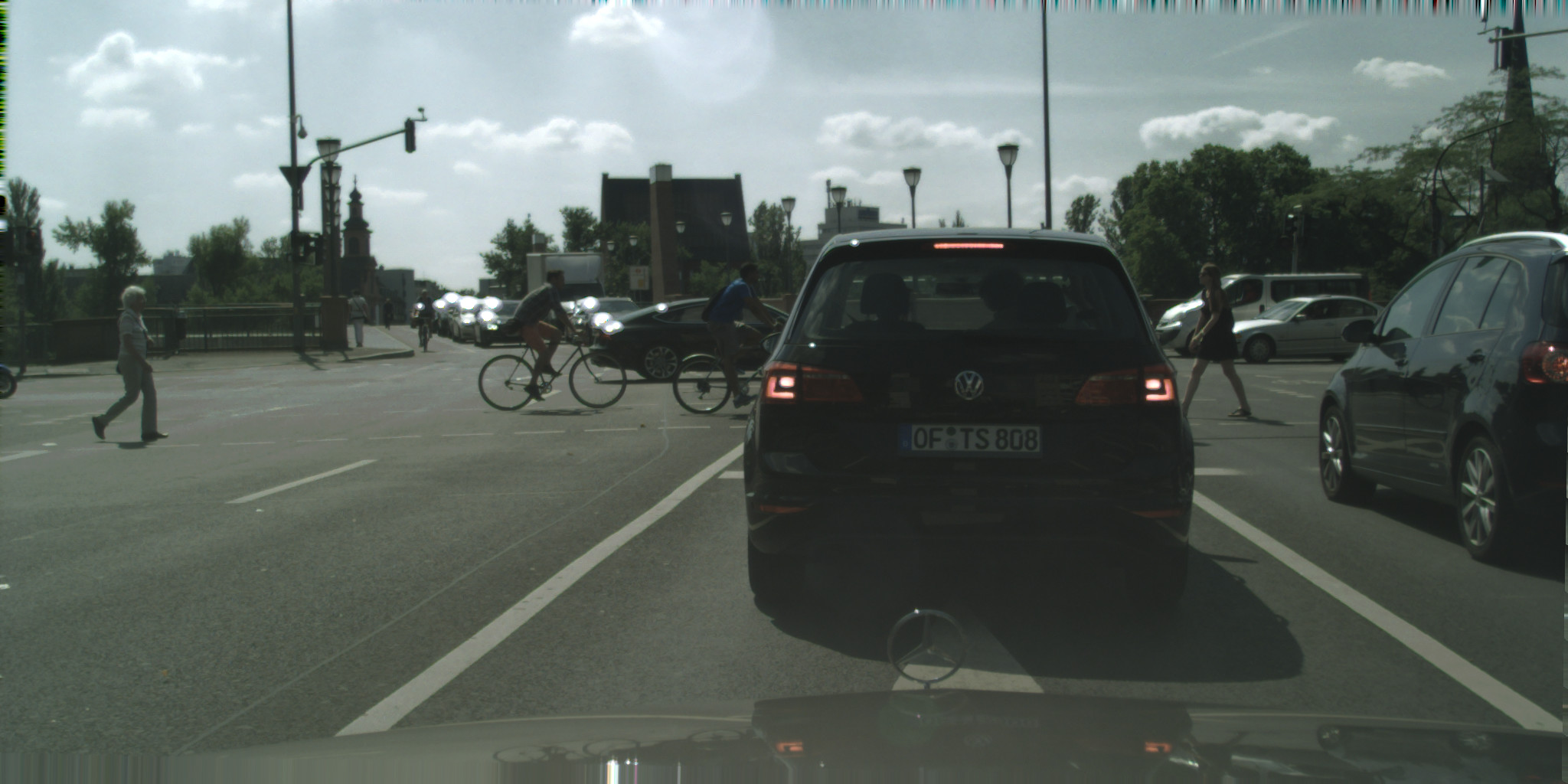}
   \includegraphics[width=0.32\linewidth]{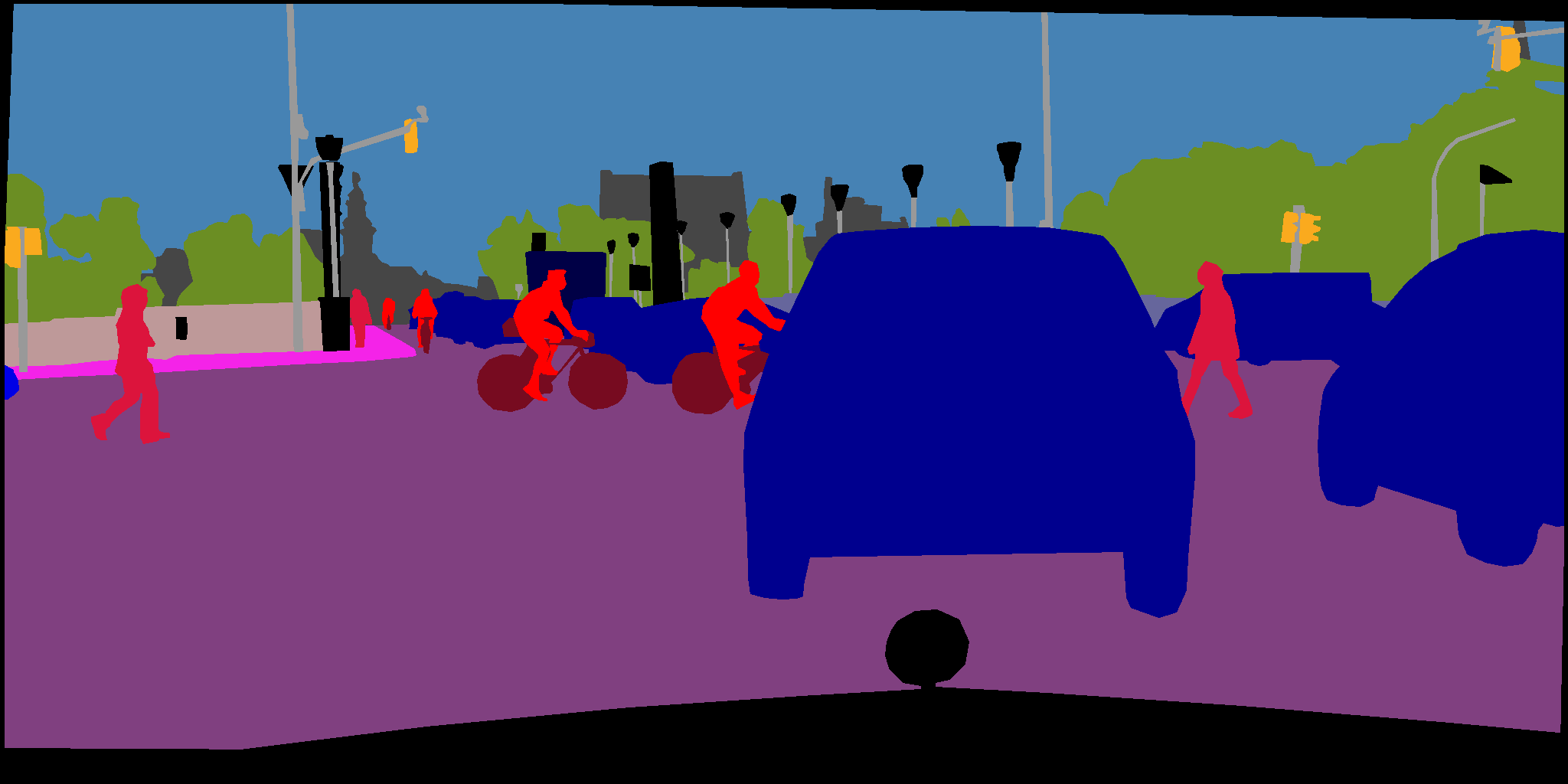}
   \includegraphics[width=0.32\linewidth]{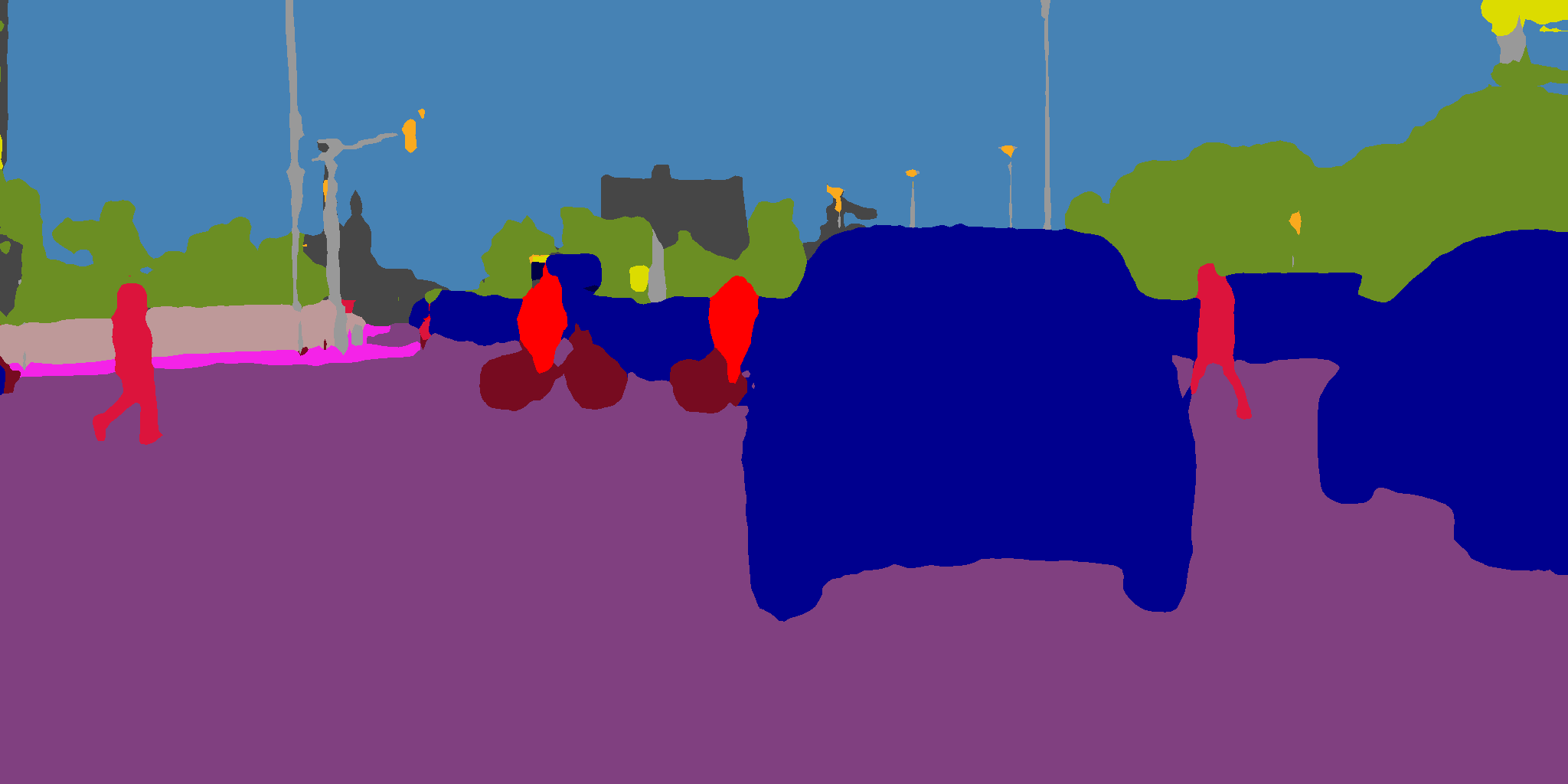}
   \\
   \includegraphics[width=0.32\linewidth]{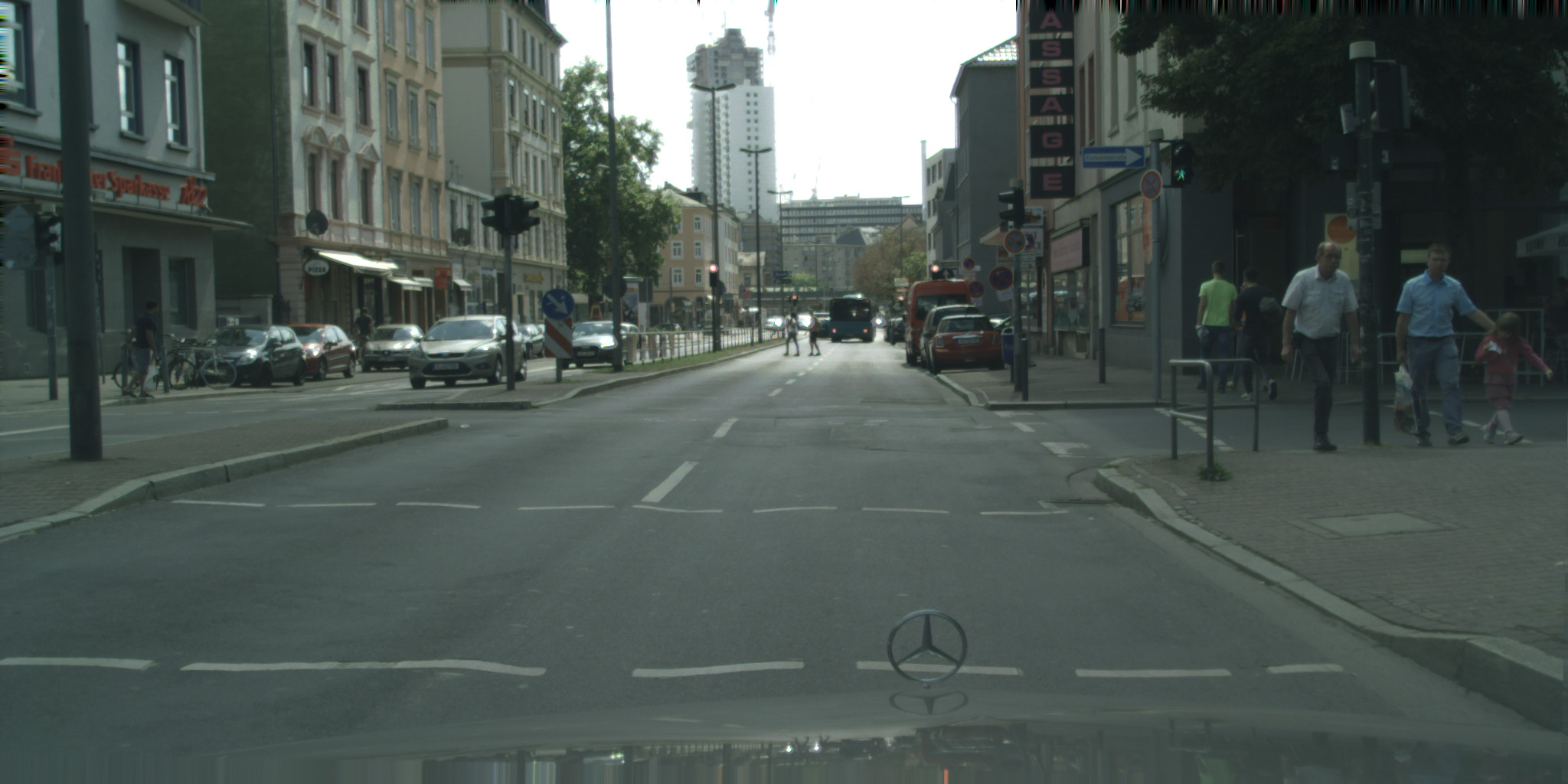}
   \includegraphics[width=0.32\linewidth]{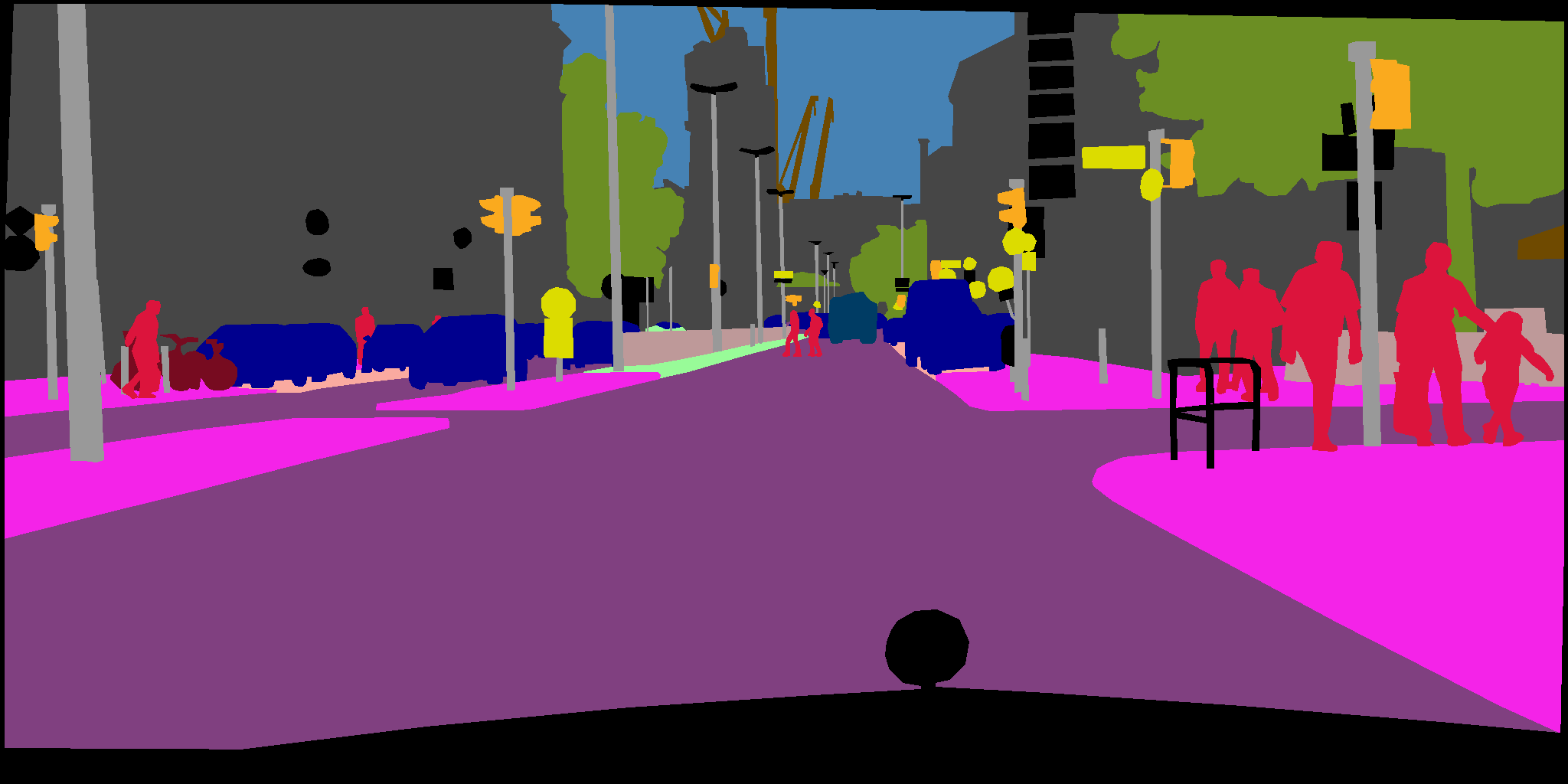}
   \includegraphics[width=0.32\linewidth]{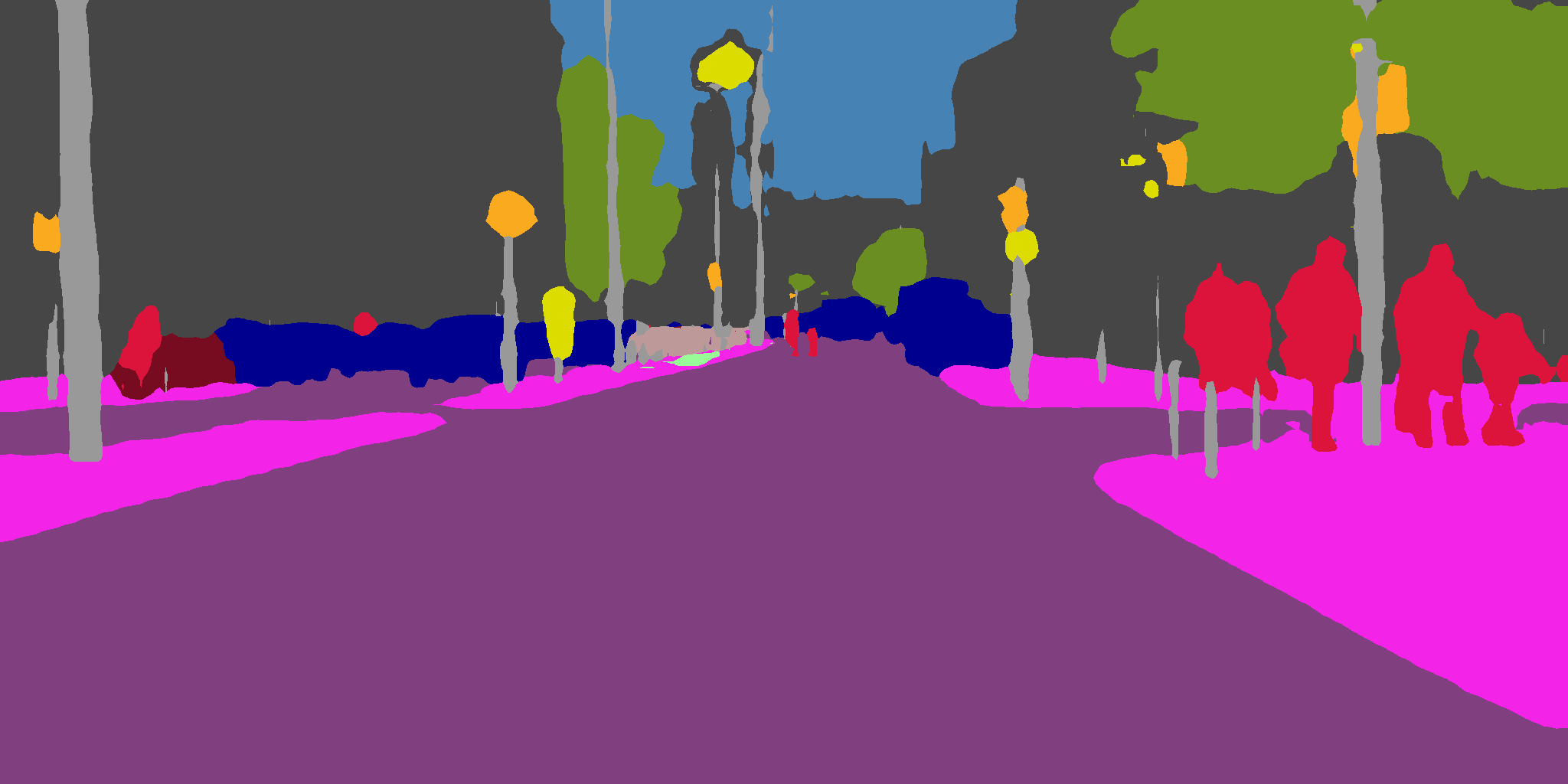}
   \\
    \includegraphics[width=0.32\linewidth]{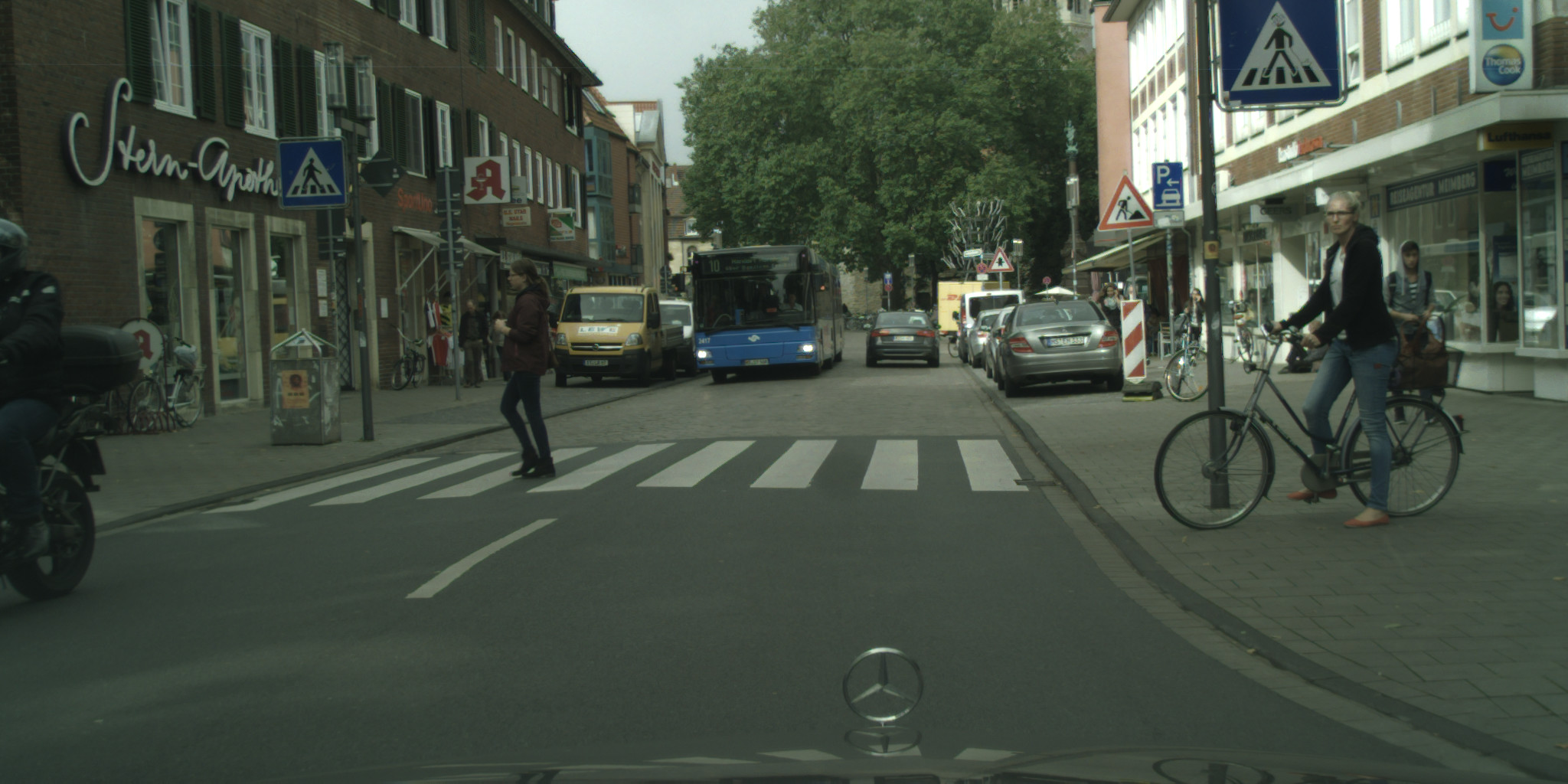}
   \includegraphics[width=0.32\linewidth]{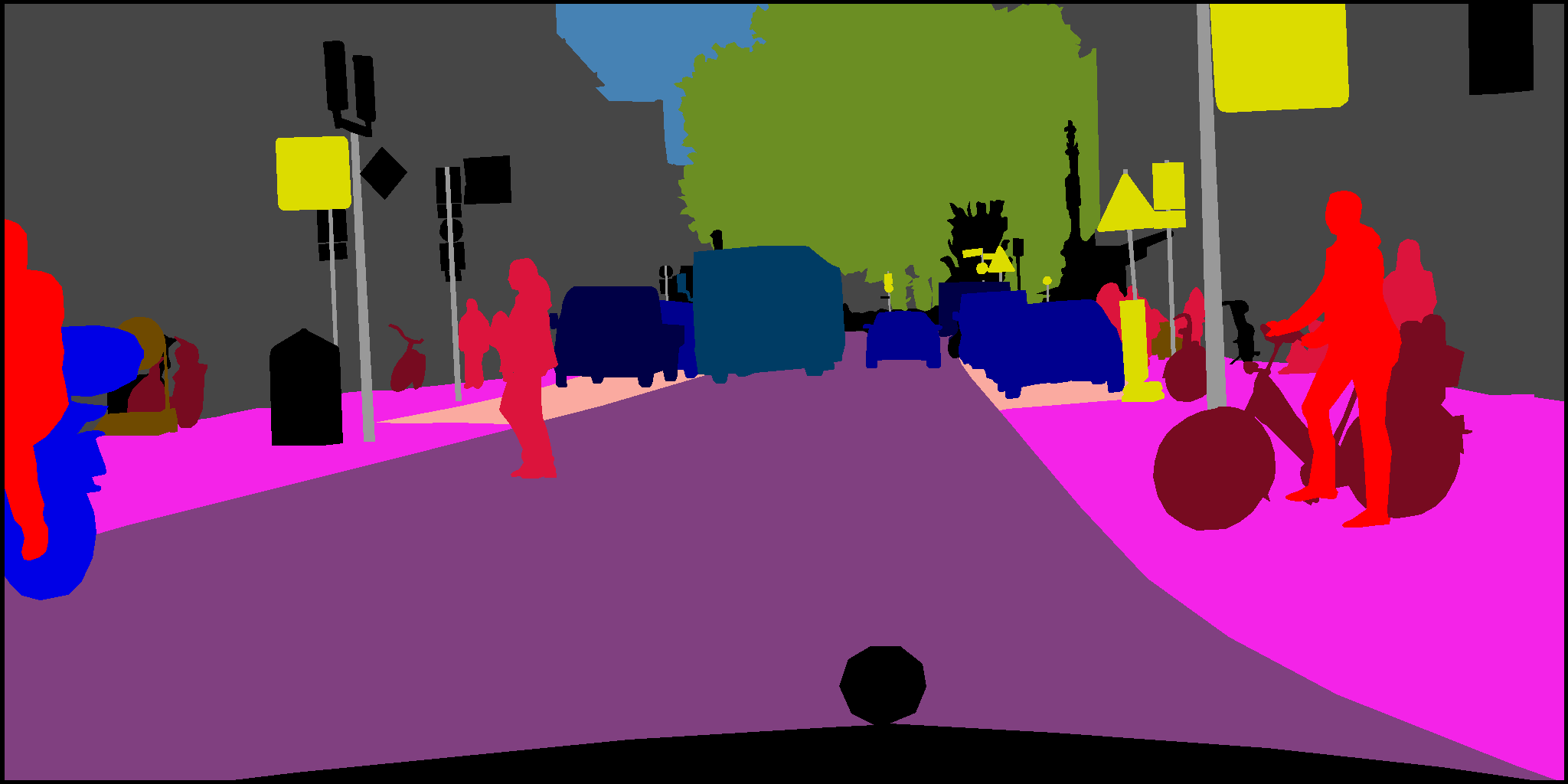}
   \includegraphics[width=0.32\linewidth]{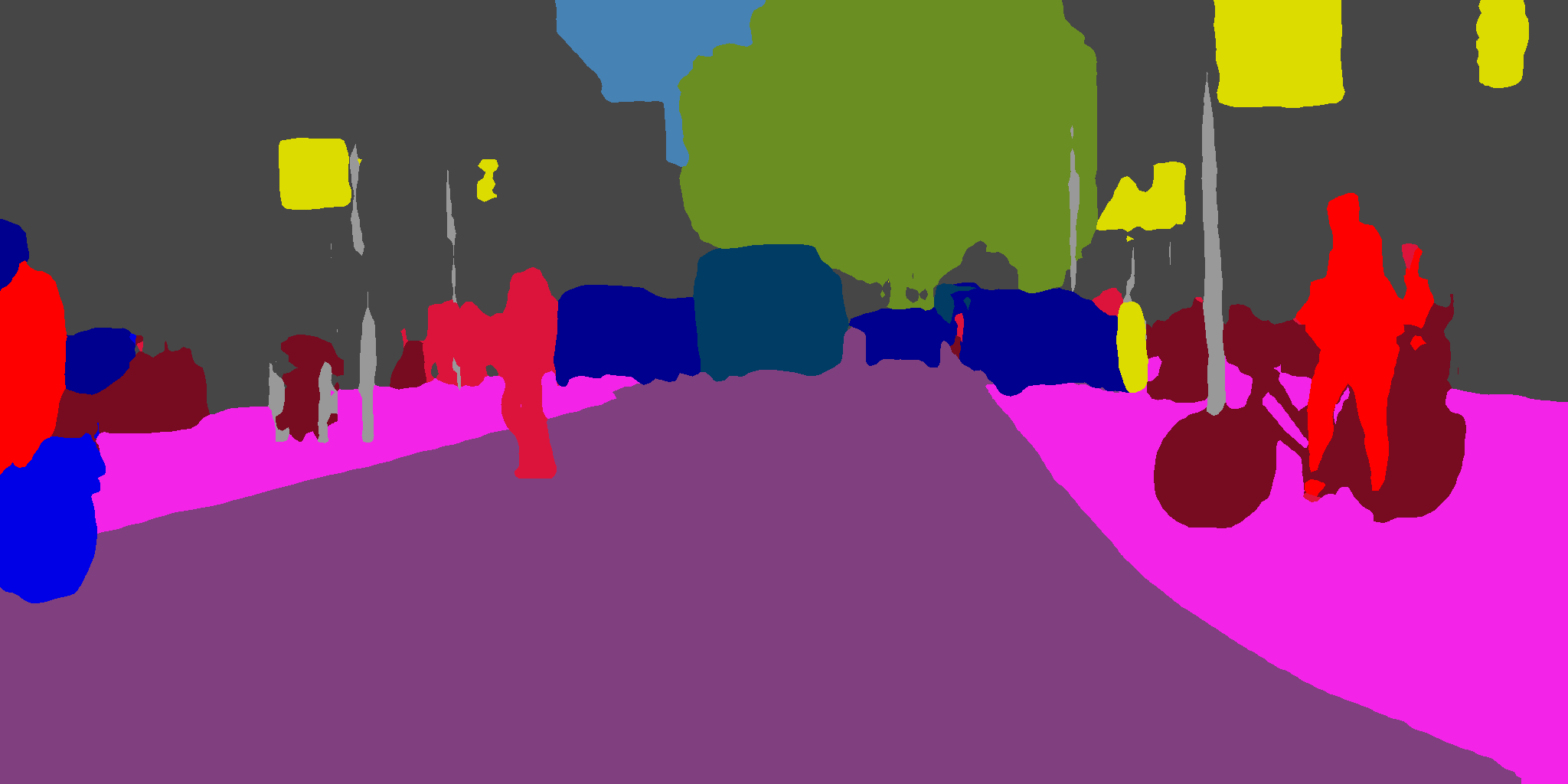}
   \\
    \includegraphics[width=0.32\linewidth]{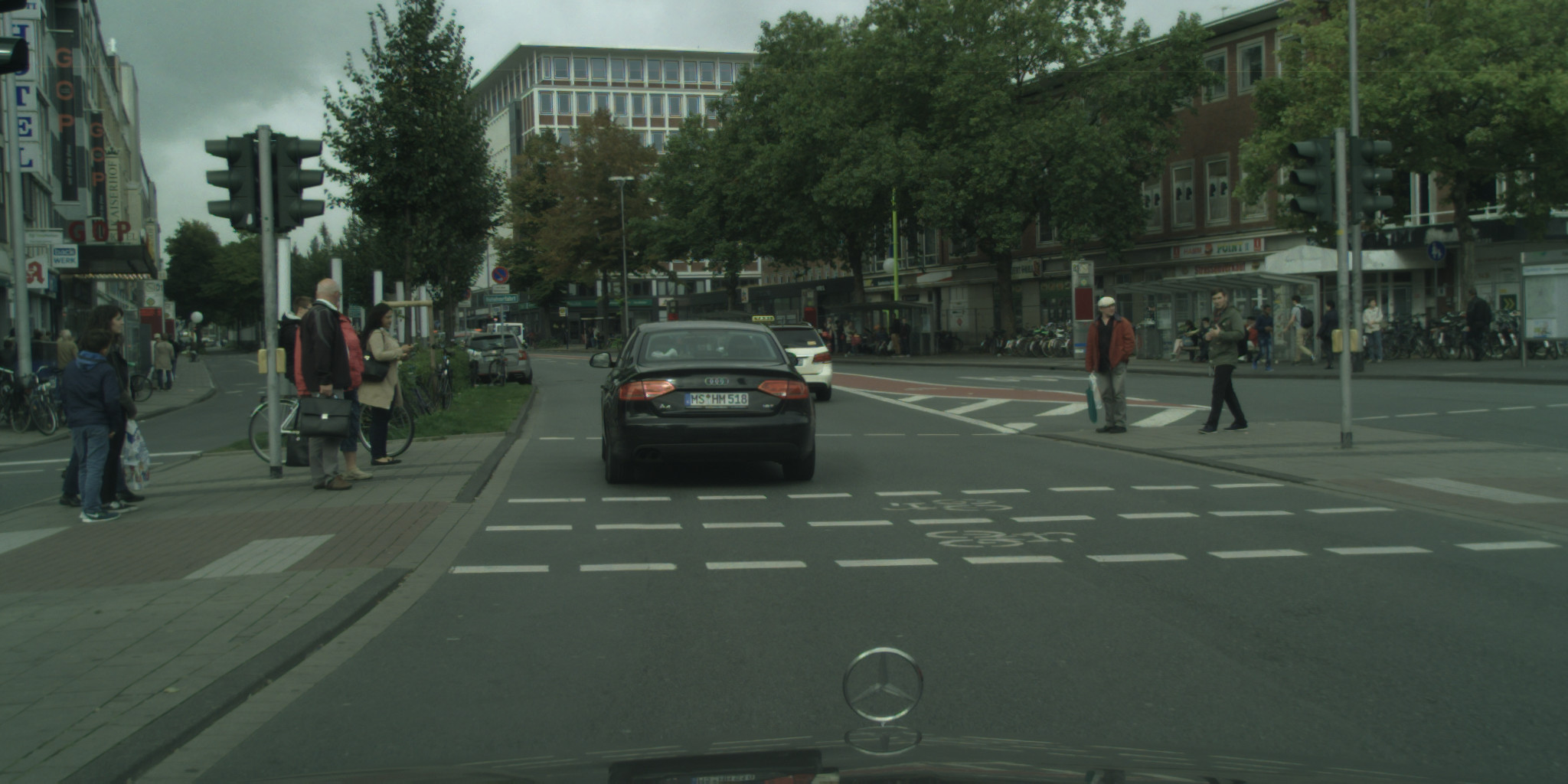}
   \includegraphics[width=0.32\linewidth]{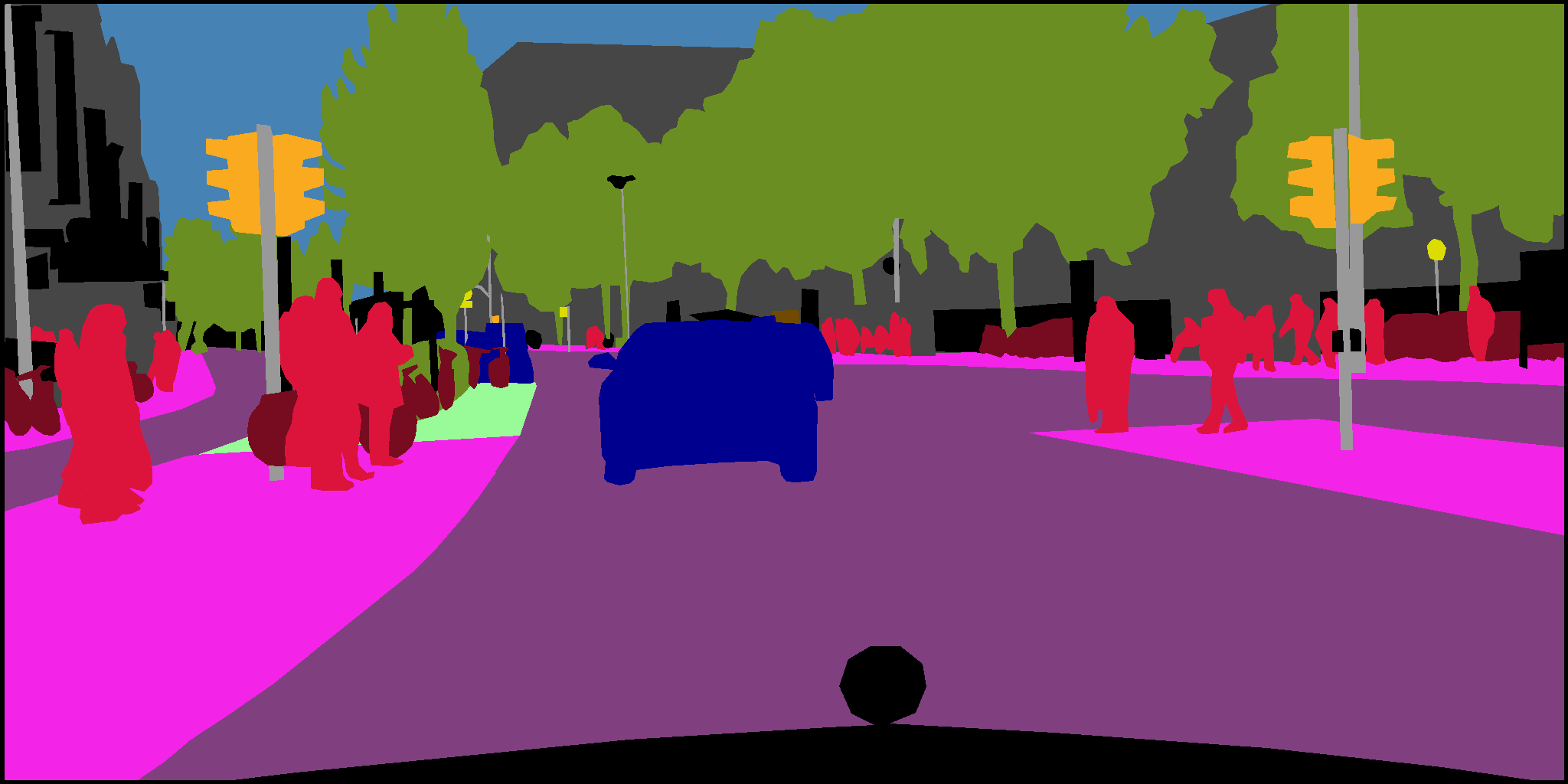}
   \includegraphics[width=0.32\linewidth]{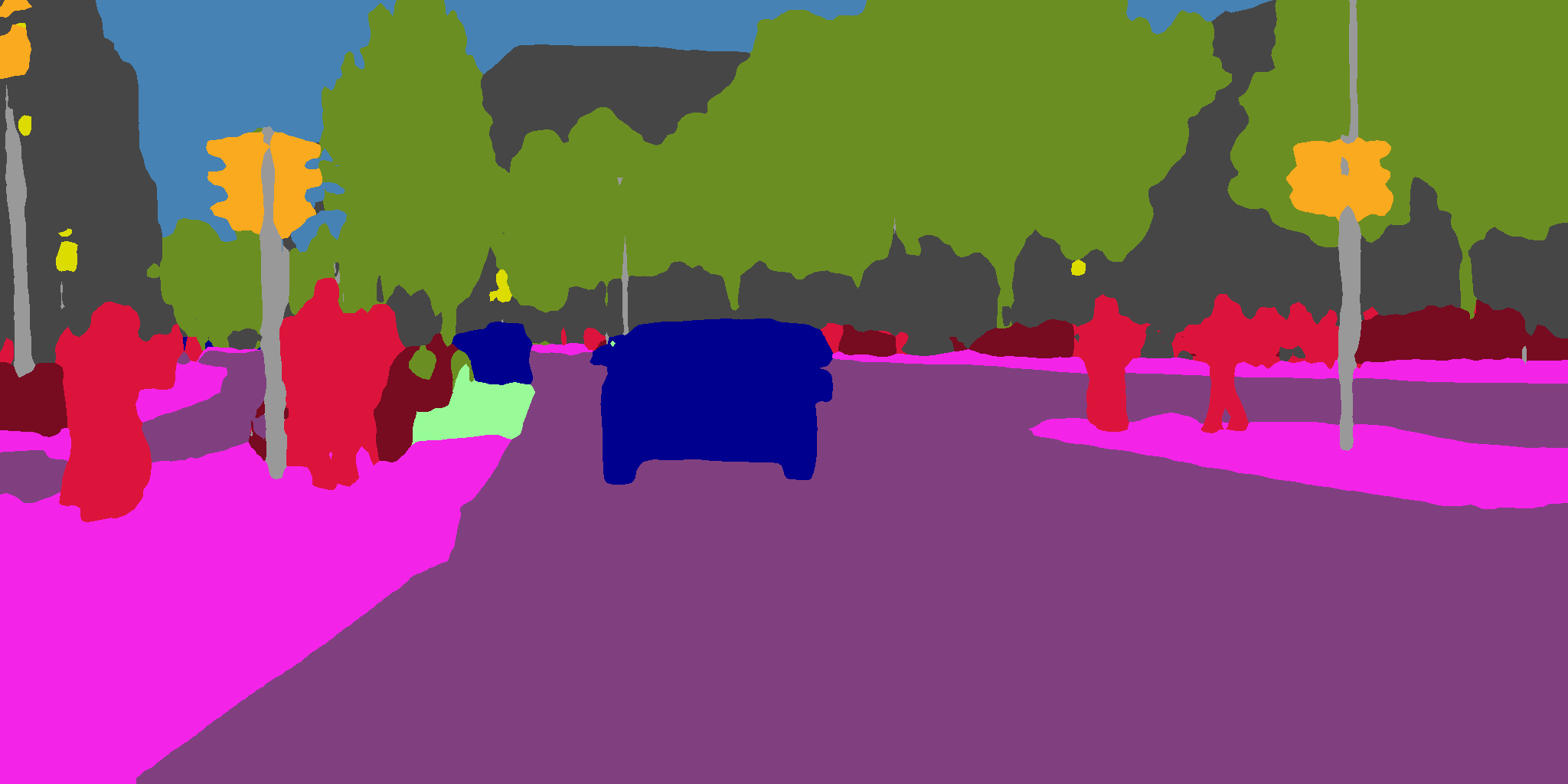}
   \\
    \includegraphics[width=0.32\linewidth]{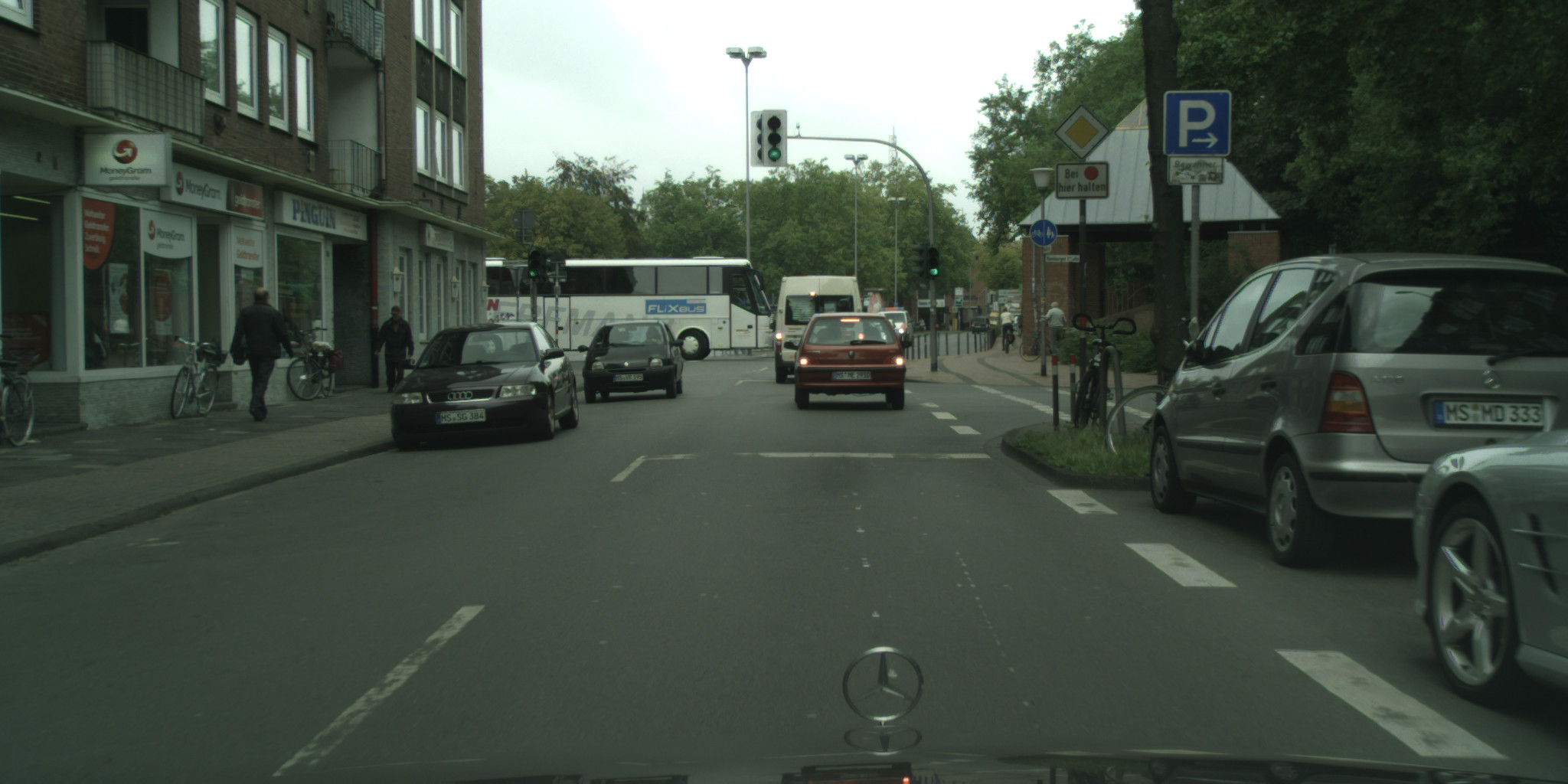}
   \includegraphics[width=0.32\linewidth]{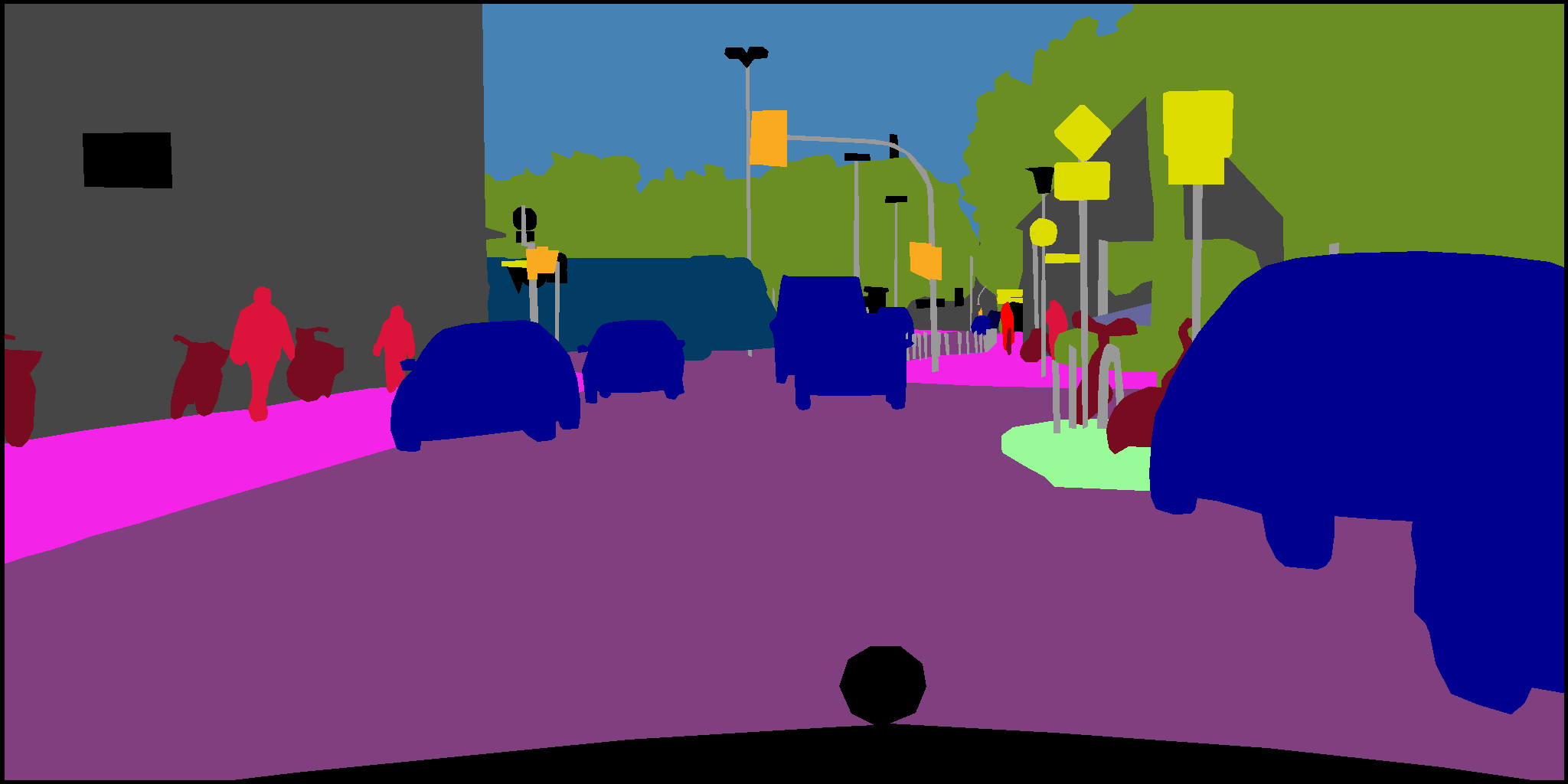}
   \includegraphics[width=0.32\linewidth]{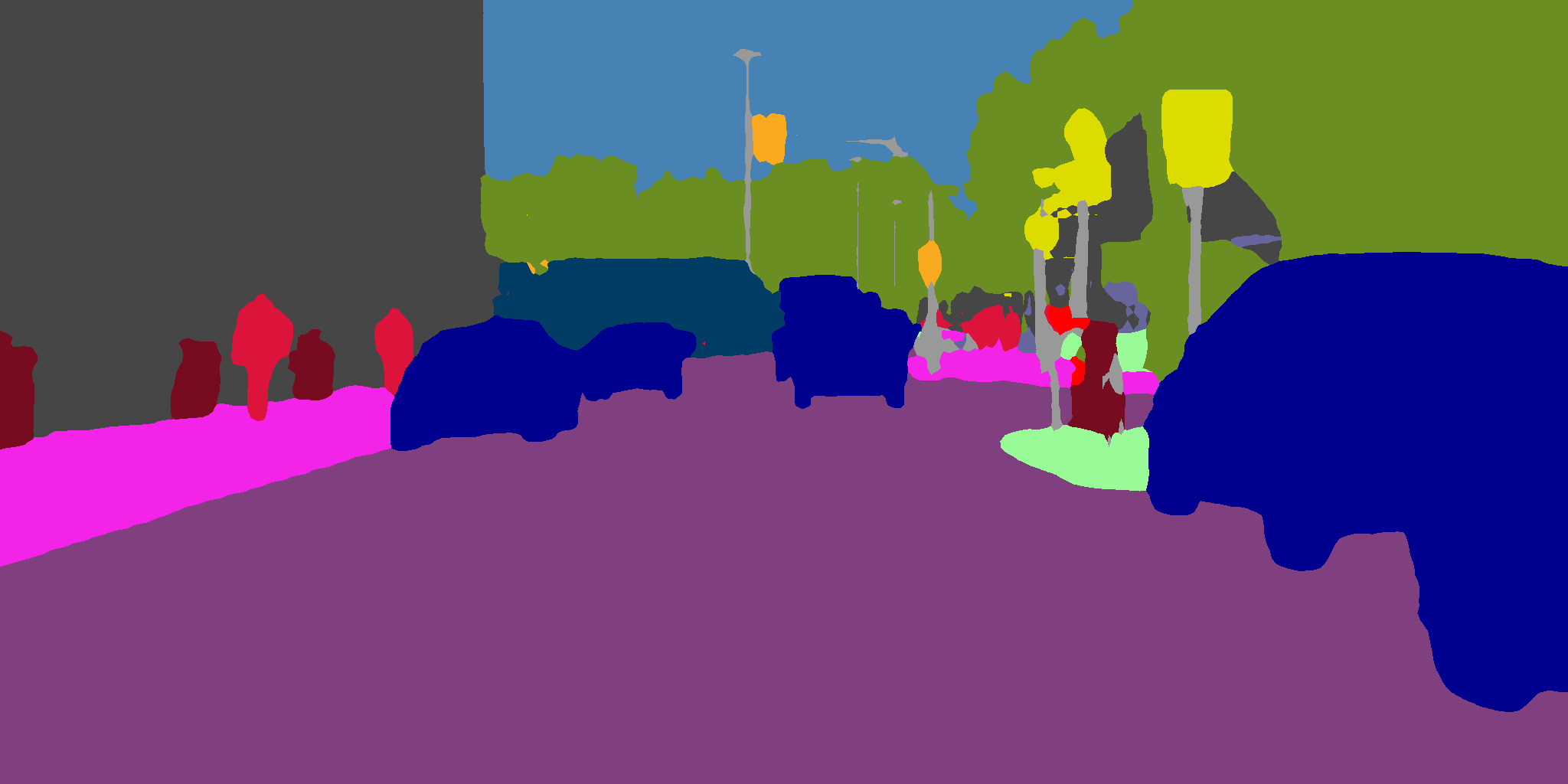}
\end{center}
   \caption{Qualitative results of Fast-SCNN on Cityscapes \cite{cityscaples2016} validation set. First column: input RGB images; second column: ground truth labels; and last column: Fast-SCNN outputs. Fast-SCNN obtains 68.0\% class level mIoU and 84.7\% category level mIoU.}
\label{fig:fast-scnn-results}
\end{figure*}

\subsection{Lower Input Resolution}
\begin{table}[tb]
\begin{center}
%\scalebox{0.95}{%
\begin{tabular}{l c c}
\hline
\textbf{Input Size} & \textbf{Class} & \textbf{FPS} \\
\hline
	$1024\times2048$ & 68.0 & 123.5 \\ %\hline
	$512\times1024$ & 62.8 & 285.8 \\ %\hline
	$256\times512$ & 51.9 & 485.4 \\ \hline
\end{tabular}
%}
\end{center}
\caption{Runtime and accuracy of Fast-SCNN at different input resolutions on Cityscapes' test set \cite{cityscaples2016}.}
\label{tbl:comparision-diff-resolutions}
\end{table}
Since we are interested in embedded devices that may not have full resolution input, or access to powerful GPUs, we conclude our evaluation with the study of performance at half, and quarter input resolutions (Table~\ref{tbl:comparision-diff-resolutions}).

At quarter resolution, Fast-SCNN achieves 51.9\% accuracy at 485.4 fps, which significantly improves on (anonymous) MiniNet with 40.7\% mIoU at 250 fps \cite{cityscaples2016}. At half resolution, a competitive 62.8\% mIoU at 285.8 fps is reached. We emphasize, without modification, Fast-SCNN is directly applicable to lower input resolution, making it highly suitable for embedded devices.

%------------------------------------------------------------------------
\section{Conclusions}
\label{sec:conclusions}
We propose a fast segmentation network for above real-time scene understanding. Sharing the computational cost of the multi-branch network yields run-time efficiency. In experiments our skip connection is shown beneficial for recovering the spatial details. We also demonstrate that if trained for long enough, large-scale pre-training of the model on an additional auxiliary task is not necessary for the low capacity network.

{\small
\bibliographystyle{ieee}
\bibliography{fast-scnn_final}

\begin{thebibliography}{10}\itemsep=-1pt

\bibitem{tensorflow2015}
M.~Abadi and et. al.
\newblock {TensorFlow}: Large-scale machine learning on heterogeneous systems,
  2015.

\bibitem{segnet-badrinarayanan2017}
V.~Badrinarayanan, A.~Kendall, and R.~Cipolla.
\newblock {SegNet}: {A} {Deep} {Convolutional} {Encoder}-{Decoder}
  {Architecture} for {Image} {Segmentation}.
\newblock {\em TPAMI}, 2017.

\bibitem{chen2014}
L.-C. Chen, G.~Papandreou, I.~Kokkinos, K.~Murphy, and A.~L. Yuille.
\newblock Semantic image segmentation with deep convolutional nets and fully
  connected crfs, 2014.

\bibitem{deeplab-v2-chen2016}
L.-C. Chen, G.~Papandreou, I.~Kokkinos, K.~Murphy, and A.~L. Yuille.
\newblock {DeepLab}: {Semantic} {Image} {Segmentation} with {Deep}
  {Convolutional} {Nets}, {Atrous} {Convolution}, and {Fully} {Connected}
  {CRFs}.
\newblock {\em arXiv:1606.00915 [cs]}, 2016.

\bibitem{xception-net-chollet2016}
F.~Chollet.
\newblock Xception: {Deep} {Learning} with {Depthwise} {Separable}
  {Convolutions}.
\newblock {\em arXiv:1610.02357 [cs]}, 2016.

\bibitem{cityscaples2016}
M.~Cordts, M.~Omran, S.~Ramos, T.~Rehfeld, M.~Enzweiler, R.~Benenson,
  U.~Franke, S.~Roth, and B.~Schiele.
\newblock The cityscapes dataset for semantic urban scene understanding.
\newblock In {\em CVPR}, 2016.

\bibitem{rcnn-girshick2013}
R.~Girshick, J.~Donahue, T.~Darrell, and J.~Malik.
\newblock Rich feature hierarchies for accurate object detection and semantic
  segmentation, 2013.

\bibitem{deep-compression-han2016}
S.~Han, H.~Mao, and W.~J. Dally.
\newblock Deep {Compression}: {Compressing} {Deep} {Neural} {Networks} with
  {Pruning}, {Trained} {Quantization} and {Huffman} {Coding}.
\newblock In {\em {ICLR}}, 2016.

\bibitem{resnet-he2015}
K.~He, X.~Zhang, S.~Ren, and J.~Sun.
\newblock Deep {Residual} {Learning} for {Image} {Recognition}.
\newblock {\em arXiv:1512.03385 [cs]}, 2015.

\bibitem{mobilenet-howard2017}
A.~Howard, M.~Zhu, B.~Chen, D.~Kalenichenko, W.~Wang, T.~Weyand, M.~Andreetto,
  and H.~Adam.
\newblock {MobileNets}: {Efficient} {Convolutional} {Neural} {Networks} for
  {Mobile} {Vision} {Applications}.
\newblock {\em arXiv:1704.04861 [cs]}, 2017.

\bibitem{hubara2016}
I.~Hubara, M.~Courbariaux, D.~Soudry, R.~El-Yaniv, and Y.~Bengio.
\newblock Binarized {Neural} {Networks}.
\newblock In {\em {NIPS}}. 2016.

\bibitem{batch-norm-ioffe2015}
S.~Ioffe and C.~Szegedy.
\newblock Batch {Normalization}: {Accelerating} {Deep} {Network} {Training} by
  {Reducing} {Internal} {Covariate} {Shift}.
\newblock {\em arXiv:1502.03167 [cs]}, 2015.

\bibitem{lazebnik2006}
S.~Lazebnik, C.~Schmid, and J.~Ponce.
\newblock Beyond bags of features: Spatial pyramid matching for recognizing
  natural scene categories.
\newblock In {\em CVPR}, volume~2, pages 2169--2178, 2006.

\bibitem{pruning-fliters-li2017}
H.~Li, A.~Kadav, I.~Durdanovic, H.~Samet, and H.~P. Graf.
\newblock Pruning {Filters} for {Efficient} {ConvNets}.
\newblock In {\em {ICLR}}, 2017.

\bibitem{ssd-liu2015}
W.~Liu, D.~Anguelov, D.~Erhan, C.~Szegedy, S.~Reed, C.-Y. Fu, and A.~C. Berg.
\newblock Ssd: Single shot multibox detector.
\newblock 2015.

\bibitem{lucchi2011}
A.~Lucchi, Y.~Li, X.~B. Bosch, K.~Smith, and P.~Fua.
\newblock Are spatial and global constraints really necessary for segmentation?
\newblock In {\em ICCV}, 2011.

\bibitem{gun-mazzini2018}
D.~Mazzini.
\newblock Guided {Upsampling} {Network} for {Real}-{Time} {Semantic}
  {Segmentation}.
\newblock In {\em BMVC}, 2018.

\bibitem{espnet-mehta2018}
S.~Mehta, M.~Rastegari, A.~Caspi, L.~Shapiro, and H.~Hajishirzi.
\newblock {ESPNet}: {Efficient} {Spatial} {Pyramid} of {Dilated} {Convolutions}
  for {Semantic} {Segmentation}.
\newblock {\em arXiv:1803.06815 [cs]}, 2018.

\bibitem{olah2017}
C.~Olah, A.~Mordvintsev, and L.~Schubert.
\newblock Feature visualization.
\newblock {\em Distill}, 2017.

\bibitem{enet-paszke2016}
A.~Paszke, A.~Chaurasia, S.~Kim, and E.~Culurciello.
\newblock {ENet}: {A} {Deep} {Neural} {Network} {Architecture} for
  {Real}-{Time} {Semantic} {Segmentation}.
\newblock {\em arXiv:1606.02147 [cs]}, 2016.

\bibitem{contextnet-poudel2018}
R.~Poudel, U.~Bonde, S.~Liwicki, and S.~Zach.
\newblock Contextnet: Exploring context and detail for semantic segmentation in
  real-time.
\newblock In {\em BMVC}, 2018.

\bibitem{xnornet-rastegari2016}
M.~Rastegari, V.~Ordonez, J.~Redmon, and A.~Farhadi.
\newblock {XNOR}-{Net}: {ImageNet} {Classification} {Using} {Binary}
  {Convolutional} {Neural} {Networks}.
\newblock In {\em {ECCV}}, 2016.

\bibitem{yolo-redmon2016}
J.~Redmon, S.~Divvala, R.~Girshick, and A.~Farhadi.
\newblock You only look once: Unified, real-time object detection.
\newblock In {\em CVPR}, 2016.

\bibitem{yolo9000-redmon2016}
J.~Redmon and A.~Farhadi.
\newblock Yolo9000: Better, faster, stronger, 2016.

\bibitem{erfnet-romera2018}
E.~Romera, J.~M. \'{A}lvarez, L.~M. Bergasa, and R.~Arroyo.
\newblock {ERFNet}: {Efficient} {Residual} {Factorized} {ConvNet} for
  {Real}-{Time} {Semantic} {Segmentation}.
\newblock {\em IEEE Transactions on Intelligent Transportation Systems}, 2018.

\bibitem{u-net-ronneberger2015}
O.~Ronneberger, P.~Fischer, and T.~Brox.
\newblock U-{Net}: {Convolutional} {Networks} for {Biomedical} {Image}
  {Segmentation}.
\newblock In {\em {MICCAI}}, 2015.

\bibitem{imagenet2015}
O.~Russakovsky, J.~Deng, H.~Su, J.~Krause, S.~Satheesh, S.~Ma, Z.~Huang,
  A.~Karpathy, A.~Khosla, M.~Bernstein, A.~Berg, and L.~Fei-Fei.
\newblock {ImageNet Large Scale Visual Recognition Challenge}.
\newblock {\em International Journal of Computer Vision (IJCV)}, 2015.

\bibitem{inverted-res-bottlenecks-sandler2018}
M.~Sandler, A.~Howard, M.~Zhu, A.~Zhmoginov, and L.-C. Chen.
\newblock Inverted {Residuals} and {Linear} {Bottlenecks}: {Mobile} {Networks}
  for {Classification}, {Detection} and {Segmentation}.
\newblock {\em arXiv:1801.04381 [cs]}, 2018.

\bibitem{fcn-long2016}
E.~Shelhamer, J.~Long, and T.~Darrell.
\newblock Fully convolutional networks for semantic segmentation.
\newblock {\em PAMI}, 2016.

\bibitem{depthwise-conv-sifre2014}
L.~Sifre.
\newblock {\em Rigid-motion scattering for image classification}.
\newblock PhD thesis, 2014.

\bibitem{vgg-simonyan2014}
K.~Simonyan and A.~Zisserman.
\newblock Very deep convolutional networks for large-scale image recognition.
\newblock {\em CoRR}, abs/1409.1556, 2014.

\bibitem{visin2015}
F.~Visin, M.~Ciccone, A.~Romero, K.~Kastner, K.~Cho, Y.~Bengio, M.~Matteucci,
  and A.~Courville.
\newblock Reseg: A recurrent neural network-based model for semantic
  segmentation, 2015.

\bibitem{wu2018}
S.~Wu, G.~Li, F.~Chen, and L.~Shi.
\newblock Training and {Inference} with {Integers} in {Deep} {Neural}
  {Networks}.
\newblock In {\em {ICLR}}, 2018.

\bibitem{BiSeNet-yu2018}
C.~Yu, J.~Wang, C.~Peng, C.~Gao, G.~Yu, and N.~Sang.
\newblock Bisenet: Bilateral segmentation network for real-time semantic
  segmentation.
\newblock In {\em ECCV}, 2018.

\bibitem{deconv-zeiler2014}
M.~D. Zeiler and R.~Fergus.
\newblock Visualizing and {Understanding} {Convolutional} {Networks}.
\newblock In {\em {ECCV}}, 2014.

\bibitem{icnet-zhao2017b}
H.~Zhao, X.~Qi, X.~Shen, J.~Shi, and J.~Jia.
\newblock {ICNet} for {Real}-{Time} {Semantic} {Segmentation} on
  {High}-{Resolution} {Images}.
\newblock In {\em {ECCV}}, 2018.

\bibitem{pspnet-zhao2017a}
H.~Zhao, J.~Shi, X.~Qi, X.~Wang, and J.~Jia.
\newblock Pyramid {Scene} {Parsing} {Network}.
\newblock In {\em CVPR}, 2017.

\bibitem{zheng2015}
S.~Zheng, S.~Jayasumana, B.~Romera-Paredes, V.~Vineet, Z.~Su, D.~Du, C.~Huang,
  and P.~H.~S. Torr.
\newblock Conditional random fields as recurrent neural networks.
\newblock In {\em ICCV}, December 2015.

\end{thebibliography}
}

\end{document}